\documentclass[letterpaper]{article}      % AAAI kit: DO NOT CHANGE THIS
\usepackage{aaai2027}                     % AAAI kit (no [submission] option: this is the non-anonymous version)
\usepackage{times}                        % AAAI kit: DO NOT CHANGE THIS
\usepackage{helvet}                       % AAAI kit: DO NOT CHANGE THIS
\usepackage{courier}                      % AAAI kit: DO NOT CHANGE THIS
\usepackage[hyphens]{url}                 % AAAI kit: DO NOT CHANGE THIS
\usepackage{graphicx}                     % AAAI kit: DO NOT CHANGE THIS
\usepackage{natbib}                       % AAAI kit: DO NOT CHANGE THIS AND DO NOT ADD ANY OPTIONS TO IT
\usepackage{caption}                      % AAAI kit: DO NOT CHANGE THIS AND DO NOT ADD ANY OPTIONS TO IT
\usepackage{amsmath}
\usepackage{amssymb}
\usepackage{booktabs}
\usepackage{makecell}
\usepackage{xcolor}
\usepackage{pdfpages}
\newcommand{\ind}[1]{\mathbf{1}\!\left[#1\right]}

\begin{document}

\title{GEMS-RAG: Graph-grounded Engineering Multimodal Standards Retrieval-Augmented Generation}

%% ARXIV: author block. Fill in from the author list and affiliations.
%% AAAI kit rules: superscripts in \rm, no space between the name (or its comma) and the
%% superscript, \equalcontrib after a name marks equal contribution, e-mails in roman type.
\author{
    Abdul Hannan Azad\textsuperscript{\rm 1},
    Irvin Wang\textsuperscript{\rm 2},
    Jiwan Jiang\textsuperscript{\rm 1},
    Chenchen Kuai\textsuperscript{\rm 1},
    Bofeng Cao\textsuperscript{\rm 3},\\
    Keshu Wu\textsuperscript{\rm 1},
    Yuzhou Wang\textsuperscript{\rm 1},
    Xiaopeng Li\textsuperscript{\rm 3},
    Yang Zhou\textsuperscript{\rm 1}
}
\affiliations{
    \textsuperscript{\rm 1}Texas A\&M University, College Station, USA\\
    \textsuperscript{\rm 2}A\&M Consolidated High School College Station, USA\\
    \textsuperscript{\rm 3}University of Wisconsin, Madison, USA\\
    yangzhou295@tamu.edu
}

    % \{hannan\_123, jiwan.jiang, mobility, keshuw, yuzhouw, yangzhou295\} @tamu.edu,\\
    % \{Irvinwop\} @gmail.com,\\
    % \{bcao44, xli2485\} @wisc.edu\\

%% ARXIV: the AAAI copyright line must not appear on a preprint of a paper that has not been
%% accepted. Remove this line for the camera-ready (AAAI will not publish a paper that uses it).
%\nocopyright

\maketitle

\begin{abstract}
Engineering standards are challenging knowledge sources for large language models (LLMs) because they are hierarchical, multimodal, and densely connected through figures, tables, cross-references, and normative rules. Existing retrieval-augmented generation (RAG) methods often rely on unstructured text retrieval and struggle with questions requiring joint reasoning over textual provisions, visual specifications, and document structure. We propose \textbf{GEMS-RAG} (\textbf{G}raph-grounded \textbf{E}ngineering \textbf{M}ultimodal \textbf{S}tandards Retrieval-Augmented Generation), a framework for grounded reasoning over engineering standards. GEMS-RAG combines hierarchical parsing, fine-grained visual entity extraction, multimodal knowledge graph construction, and graph-grounded hybrid retrieval. Instead of treating figures as monolithic images, it decomposes them into meaningful visual entities linked to textual provisions, engineering concepts, and regulatory rules. During inference, dense, sparse, and visual retrieval are combined with graph expansion to identify relevant evidence across text, figures, and document structure. We evaluate GEMS-RAG on the \textit{Manual on Uniform Traffic Control Devices} (MUTCD), a large engineering standard containing regulations, technical drawings, and standardized traffic-control devices. Results show that graph-grounded multimodal retrieval improves retrieval effectiveness and question-answering performance over text-only, multimodal, and graph-based RAG baselines. Beyond MUTCD, GEMS-RAG offers a general framework for reasoning over engineering codes, technical manuals, regulatory documents, and scientific standards in safety-critical applications.
\end{abstract}

%% AAAI kit: the links block goes between the abstract and the body, never inside either.
%% ARXIV: put the public repository address here.
\begin{links}
    \link{Code}{https://github.com/hannanazad/gemsrag.git}
\end{links}

\section{Introduction}

Large Language Models (LLMs) have demonstrated remarkable capabilities in language understanding, reasoning, and question answering. However, their effectiveness remains fundamentally constrained by the knowledge encoded during pretraining \citep{kuai2026independent}. Retrieval-Augmented Generation (RAG) has emerged as a powerful paradigm for grounding LLMs in external knowledge by retrieving relevant documents at inference time and incorporating them into the generation process \citep{lewis2020retrieval,lewis_retrieval-augmented_2021, asai_self-rag_2023, zhai_self-adaptive_2024}. While recent RAG systems have achieved substantial success on web documents \citep{yang2024crag}, scientific literature \citep{lala2023paperqa}, and domain knowledge bases \citep{kuai2026cyportqa}, they remain poorly suited for regulatory and engineering standards documents that exhibit rich hierarchical organization, multimodal content, and normative reasoning requirements.

Engineering standards documents constitute a particularly challenging class of knowledge sources for retrieval-augmented generation. Unlike conventional text corpora, they combine hierarchical organization, normative language, technical specifications, figures, tables, design examples, and dense cross-references. For example, the Manual on Uniform Traffic Control Devices (MUTCD) \citep{fhwa2025mutcd} contains textual regulations, engineering guidance, standardized traffic-control devices, technical drawings, sign specifications, warrants, and cross-referenced provisions. In such documents, critical knowledge is rarely contained in a single passage; instead, correct interpretation often requires connecting textual rules with visual specifications, document structure, and regulatory context \citep{doris2025designqa, shi2026trust}.

Conventional RAG systems face three key limitations when applied to engineering standards documents such as the MUTCD. First, many retrieval pipelines represent documents as collections of independent text chunks, which can weaken the hierarchical context that determines the scope and interpretation of regulatory provisions. Second, even when multimodal content is considered, figures and diagrams are often indexed only at the page, caption, or image level, limiting access to the fine-grained visual semantics embedded in engineering symbols, standardized devices, and technical drawings. Third, standard dense or sparse retrieval does not explicitly model cross-references, regulatory dependencies, or relationships among sections, figures, tables, devices, and rule types. As a result, existing systems may produce incomplete, decontextualized, or inconsistent answers when standards-compliant reasoning requires joint evidence from text, visuals, and document structure.

To address these limitations, we propose \textbf{GEMS-RAG}, a \textbf{G}raph-grounded \textbf{E}ngineering \textbf{M}ultimodal \textbf{S}tandards Retrieval-Augmented Generation framework for engineering standards documents which combines hierarchical document parsing, multimodal figure decomposition, dense and sparse retrieval, and graph-based knowledge organization within a unified retrieval architecture. Instead of treating figures as auxiliary content, we explicitly extract and index individual visual entities, such as traffic signs, plaques, and signal illustrations, and connect them to their corresponding textual provisions through a document knowledge graph. The resulting system enables retrieval over both textual and visual semantics while preserving the structural relationships that govern interpretation of the standard.

More specifically, we construct a multimodal knowledge graph that captures relationships among sections, figures, traffic control devices, regulatory concepts, and normative statements. During retrieval, graph-based expansion supplements traditional dense retrieval by identifying structurally related content, allowing the system to recover relevant provisions even when they are distributed across multiple sections and modalities. Retrieved textual and visual evidence are then jointly provided to a vision-language reasoning model for grounded response generation.

The contributions of this paper are threefold:
\begin{itemize}
\item We introduce a standards-aware hierarchical multimodal framework that represents engineering standards as structured collections of text, figures, tables, visual entities, cross-references, and rule types rather than isolated chunks.

\item We propose a graph-grounded hybrid retrieval mechanism integrating hierarchical parsing, textual and visual retrieval, graph expansion, and rule-aware ranking to support reasoning across sections, devices, figures, and normative relationships.

\item We develop a multimodal RAG benchmark and system for the MUTCD, as an example, showing that fine-grained visual grounding, preserved document structure, and normative rule modeling improve retrieval and question answering over complex engineering standards.
\end{itemize}

\begin{figure*}[t]
    \centering
    \includegraphics[width=0.9\textwidth]{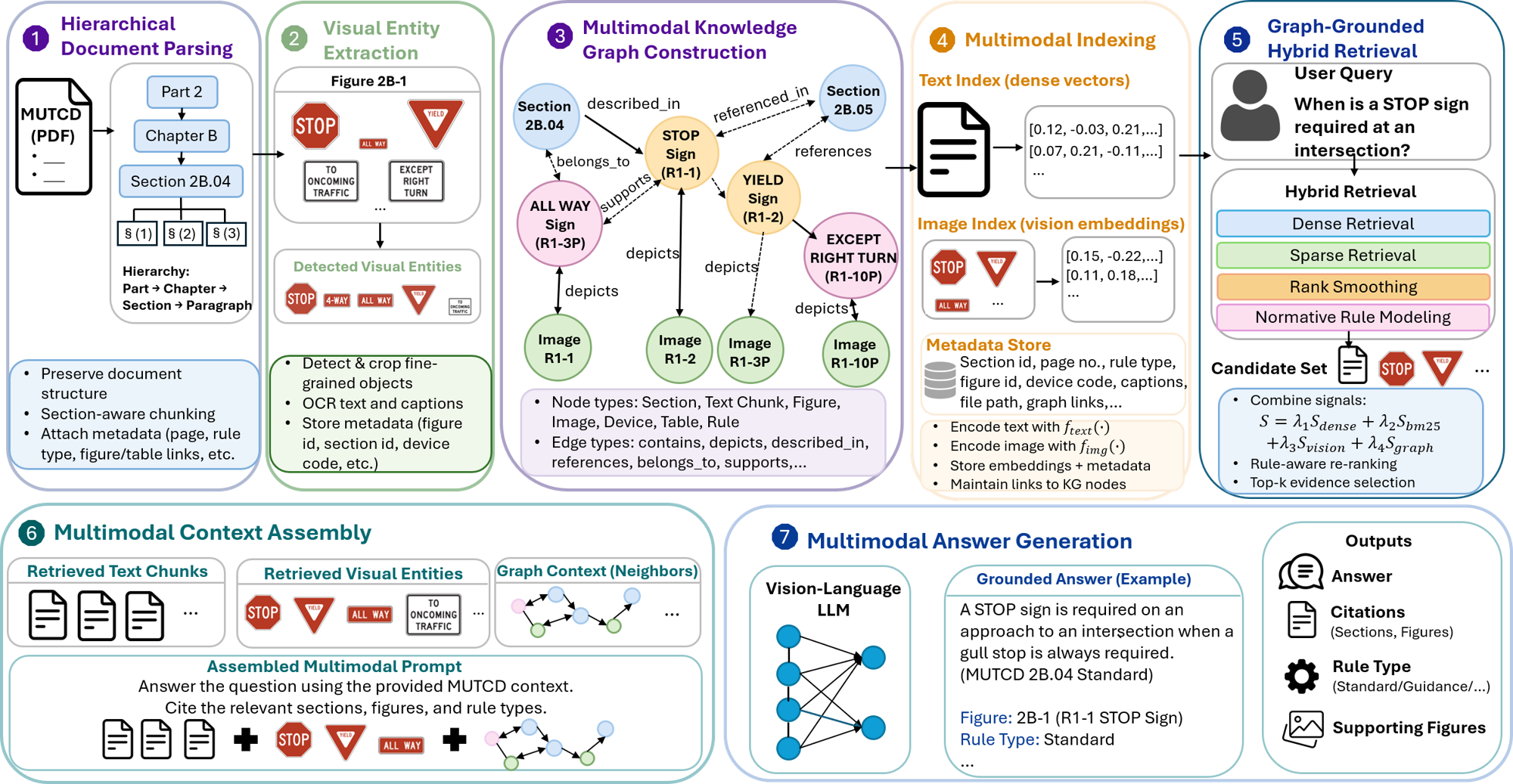}
    \caption{Framework of GEMS-RAG}
    \label{fig:methodology}
\end{figure*}

\section{Related Work}

\subsection{Retrieval-Augmented Generation for Domain-Specific Knowledge}

RAG has become a widely adopted approach for grounding large language models in external knowledge. By retrieving relevant evidence at inference time, RAG reduces reliance on parametric memory and improves factuality in question answering over open-domain documents, scientific literature, and domain-specific knowledge bases \citep{karpukhin_dense_2020}. Recent advances have further improved retrieval quality through hybrid dense--sparse retrieval, reranking, iterative retrieval, self-reflection, and adaptive retrieval planning \citep{asai_self-rag_2023, wang_learning_2024,zhai_self-adaptive_2024}. These methods have demonstrated strong potential for knowledge-intensive reasoning tasks.
Most existing RAG systems assume that knowledge can be represented as independent textual chunks. This assumption is inadequate for engineering standards, where meaning is distributed across section hierarchies, cross-references, tables, figures, captions, device drawings, and normative statements. A locally relevant passage may therefore be insufficient without its structural location, visual specifications, and related provisions. Consequently, text-centered RAG may retrieve relevant fragments yet still produce noncompliant answers. This limitation motivates a standards-aware architecture that preserves document hierarchy and regulatory context throughout retrieval and generation.

\subsection{Multimodal Understanding of Engineering Standards Documents}

Multimodal retrieval and vision-language models \citep{mathew2021docvqa,xia2025mmed,li2025pixels,kuai2026cyportqa} provide an important foundation for reasoning over visual and textual information. Shared image--text embedding models and recent multimodal foundation models have shown strong capabilities in visual question answering, document understanding, chart interpretation, and image-grounded generation. These developments have motivated multimodal RAG systems that retrieve images, tables, document pages, or visual regions alongside textual evidence \citep{yu2025visrag}.
Engineering standards present a specialized form of multimodality. Their visuals often encode formal concepts, standardized devices, geometric layouts, operational constraints, and regulatory specifications. For example, traffic signs, plaques, lane-use arrows, signal faces, and pavement markings are regulated entities with defined meanings, usage conditions, and governing provisions. Treating a figure as a monolithic image or relying only on its caption therefore obscures the fine-grained semantics required for grounded reasoning. Standards-centric multimodal RAG must explicitly extract, index, and link visual entities to their textual and regulatory context.

\subsection{Graph-Grounded Retrieval and Structural Reasoning}

Knowledge graph-augmented retrieval offers a natural mechanism for representing relationships among entities, concepts, documents, and rules \citep{procko2024graph,park2026m,hsiao2026megarag,luo2026gfm, zhu2025knowledge,wu2026knowledge}. Graph-based RAG methods can expand retrieved evidence through neighboring nodes, connect semantically related concepts, and support multi-hop reasoning beyond lexical similarity. These capabilities are especially relevant for engineering standards, where correct interpretation often depends on relationships among sections, figures, tables, definitions, cross-references, and rule types. Nevertheless, most existing graph-RAG approaches build graphs from textual entities, citations, or domain ontologies. They rarely treat engineering visuals as first-class entities or jointly represent visual objects, document hierarchy, regulatory concepts, and normative statements. Likewise, document intelligence methods can parse layouts, tables, and figures, but usually serve only as preprocessing rather than supporting graph-grounded retrieval and reasoning.

To address these limitations, this paper proposes GEMS-RAG, a graph-grounded multimodal RAG framework for engineering standards. GEMS-RAG preserves document hierarchy, decomposes figures into fine-grained visual entities, and constructs a multimodal knowledge graph linking sections, figures, traffic-control devices, regulatory concepts, and normative statements. By combining dense, sparse, and visual retrieval with graph expansion, it supports grounded reasoning across the textual, visual, and structural semantics of engineering standards. As shown in Figure~\ref{fig:methodology}, GEMS-RAG parses standards by their native hierarchy, extracts captioned figures and tables, links textual and visual elements in a knowledge graph, retrieves multimodal evidence, and adaptively gates visual evidence per query.

\section{Methodology}
\label{sec:methodology}

\subsection{Problem Formulation}
\label{subsec:problem_formulation}

Let $\mathcal{D} = (d_1,d_2,\ldots,d_N)$
denote a standards document containing textual passages, figures, tables, and cross-references. We represent each document element as $c_i = (x_i,v_i,m_i)$,
where \(x_i\) denotes the textual content, \(v_i\) denotes the visual content, and \(m_i\) contains structural metadata, including the page number, section identifier, rule type, and figure or table identifier. Given a user query \(q\), the system retrieves a supporting evidence set
\begin{equation}
\mathcal{R}(q)
=
\left\{
c_{i_1},c_{i_2},\ldots,c_{i_k}
\right\},
\label{eq:retrieved_evidence}
\end{equation}
and generates an answer
\begin{equation}
a = f\bigl(q,\mathcal{R}(q)\bigr).
\label{eq:answer_generation_general}
\end{equation}
The generated answer must ground its claims in the retrieved evidence and cite identifiers that resolve to valid sections, figures, or tables in the underlying document. Visual elements are included only when the query requires visual evidence.

\subsection{Hierarchical Document Parsing}
\label{subsec:hierarchical_parsing}

Many engineering standards organize their content into parts, chapters, sections, and typed paragraphs. GEMS-RAG preserves this native structure rather than treating the document as an unstructured sequence of pages. Each textual chunk is assigned a hierarchy label
\begin{equation}
h_i = (p_i,c_i,s_i,r_i),
\label{eq:hierarchy_label}
\end{equation}
where \(p_i\), \(c_i\), \(s_i\) and \(r_i\) denote the part, chapter, section, and normative rule type of chunk \(i\), respectively. Text is segmented by paragraph within each section, with each chunk retaining its heading, page number, and related figure, table, and section references. The parser also normalizes identifier variants to ensure consistent cross-reference matching.

\subsection{Caption-Anchored Figure and Table Extraction}
\label{subsec:caption_extraction}

GEMS-RAG uses detected captions as spatial anchors to extract figures and tables from each page. For a caption \(b\) on a page of height \(H\), the corresponding visual region is defined by
\begin{equation}
\operatorname{region}(b)
=
\left[
y_{\mathrm{bot}}(b)+\delta_1,\;
\min\left(
y_{\mathrm{top}}(b'),\;
H-\delta_2
\right)
\right],
\label{eq:caption_crop}
\end{equation}
where \(b'\) denotes the next caption below \(b\), $y_{\mathrm{bot}}(b)$ and $y_{\mathrm{top}}(b')$ define the relevant caption boundaries, and $\delta_1$ and $\delta_2$ provide small spatial margins.

The extraction procedure adapts to variations in page layout, including multi-column pages, horizontally aligned visuals, and figures distributed across multiple panels or pages. Spatial cues from caption position, column boundaries, and neighboring document elements are used to isolate the corresponding visual content. Visual components that share the same document identifier are subsequently consolidated into a single canonical entity during graph construction.

A secondary recovery procedure is applied when an identified figure or table is not successfully extracted during the initial pass. This procedure relaxes the spatial constraints to improve coverage of small or irregularly formatted visual elements. Extraction completeness is verified through a document-wide consistency check between referenced visual identifiers and the figures and tables recovered by the pipeline. Identifiers corresponding to external documents are excluded from this comparison. Any unresolved references are flagged for additional extraction or review before ingestion is completed.

\subsection{Multimodal Knowledge Graph Construction}
\label{subsec:knowledge_graph}

We organize the extracted document elements in a multimodal knowledge graph $\mathcal{G} = (\mathcal{V},\mathcal{E})$.
The node set represents the document hierarchy, textual chunks, figures, tables, and traffic control devices:
\begin{equation}
\mathcal{V}
=
\mathcal{V}_{\mathrm{part}}
\cup
\mathcal{V}_{\mathrm{chap}}
\cup
\mathcal{V}_{\mathrm{sec}}
\cup
\mathcal{V}_{\mathrm{chunk}}
\cup
\mathcal{V}_{\mathrm{fig}}
\cup
\mathcal{V}_{\mathrm{tab}}
\cup
\mathcal{V}_{\mathrm{dev}}.
\label{eq:node_set}
\end{equation}
A single canonical node is created for each figure identifier and stores the images associated with all of its sheets. The edge-type set is
\begin{equation}
\begin{aligned}
\mathcal{E}
=
\{&
\texttt{contains},
\texttt{cites},
\texttt{mentions},
\texttt{depicts},\\
&\texttt{defines},
\texttt{anchored\_in},
\texttt{cited\_in},\\
&
\texttt{belongs\_to\_chapter}
\}.
\end{aligned}
\label{eq:edge_types}
\end{equation}
The \texttt{cites} relation includes chunk-to-figure, chunk-to-table, and chunk-to-section references extracted from the chunk text.

Figures are associated with sections using three complementary forms of evidence. Let \(\mathcal{C}(f)\) denote the set of sections whose chunks cite figure \(f\), and let \(\operatorname{chap}(f)\) denote the chapter encoded in the identifier of \(f\). The primary anchor section is defined as
\begin{equation}
\operatorname{anchor}(f)
=
\underset{
s\in
\mathcal{C}(f)\cap\operatorname{chap}(f)
}{
\arg\max
}
\;
\left|
\operatorname{cites}(s,f)
\right|,
\label{eq:figure_anchor}
\end{equation}
where \(\left|\operatorname{cites}(s,f)\right|\) is the number of citations to figure \(f\) within section \(s\). If no citing section belongs to the chapter encoded in the figure identifier, the system falls back to the section with the largest number of citations to that figure.
Each figure is linked to its primary section, referencing sections, parent chapter, and depicted traffic control devices or concepts. These relations preserve the document hierarchy and each visual element's semantic role in the knowledge graph.

\subsection{Multimodal Indexing}
\label{subsec:multimodal_indexing}

Textual and visual evidence are encoded using modality-specific representation models. For each textual chunk \(x_i\), we compute a dense embedding and a learned sparse representation:
\begin{equation}
\mathbf{e}_i = f_{\mathrm{text}}(x_i),
\qquad
\mathbf{z}_i = f_{\mathrm{sparse}}(x_i),
\label{eq:text_embeddings}
\end{equation}
where both representations are produced using BGE-M3.

For a figure or table crop \(I_j\), we compute a token-level multi-vector visual representation:
\begin{equation}
\mathbf{E}_j
=
f_{\mathrm{img}}(I_j)
\in
\mathbb{R}^{n_j\times d},
\label{eq:image_embeddings}
\end{equation}
where \(n_j\) denotes the number of visual tokens and \(d\) denotes the embedding dimension. We use ColQwen2 \citep{faysse2025colpali} to encode both individual visual crops and full-page renderings. The resulting representations are stored in vector indexes, while document metadata and cross-modal relations remain in the knowledge graph.

\subsection{Graph-Grounded Hybrid Retrieval}
\label{subsec:hybrid_retrieval}

For textual retrieval, dense and sparse rankings are combined using reciprocal rank fusion:
\begin{equation}
S_{\mathrm{RRF}}(q,c_i)
=
\sum_{
r\in
\{\mathrm{dense},\mathrm{sparse}\}
}
\frac{1}{
\kappa+\operatorname{rank}_{r}(c_i)
},
\label{eq:rrf_score}
\end{equation}
where \(\kappa\) is a rank-smoothing constant. The dense ranking is induced by the cosine similarity $\cos\left(\mathbf{e}_q,\mathbf{e}_i\right)$, where \(\mathbf{e}_q\) is the dense representation of query \(q\).

Explicit figure, table, section, and device identifiers in the query are matched to a set of graph nodes \(\mathcal{V}_q\). Each retrieved candidate then receives a graph-proximity score
\begin{equation}
S_{\mathrm{graph}}(q,c_i)
=
\max_{v\in\mathcal{V}_q}
\frac{1}{
1+d_{\mathcal{G}}(v,c_i)
},
\label{eq:graph_score}
\end{equation}
where \(d_{\mathcal{G}}(\cdot,\cdot)\) denotes the shortest-path distance in \(\mathcal{G}\).

\subsection{Normative Rule Modeling}
\label{subsec:normative_modeling}

The MUTCD expresses provisions at four levels of normative authority. We assign each textual chunk a rule type
\begin{equation}
r_i
\in
\{
\textsc{Standard},
\textsc{Guidance},
\textsc{Option},
\textsc{Support}
\},
\label{eq:rule_types}
\end{equation}
and define a rule-type weight \(w(r_i)\) satisfying
\begin{equation}
\begin{aligned}
w(\textsc{Standard})
&>
w(\textsc{Guidance})\\
&>
w(\textsc{Option})
>
w(\textsc{Support}).
\end{aligned}
\label{eq:rule_weight_order}
\end{equation}

We also introduce a hierarchical relevance prior
\[
s_{\mathrm{hier}}(q,c_i)\in[0,1],
\]
which rewards candidates whose part, chapter, or section is explicitly referenced in the query. The fused first-stage retrieval score is
\begin{equation}
\begin{aligned}
S(q,c_i)
={}&
\lambda_1 S_{\mathrm{RRF}}(q,c_i)
+
\lambda_2 S_{\mathrm{graph}}(q,c_i)\\
&+
\lambda_3\bigl(w(r_i)-1\bigr)
+
\lambda_4 s_{\mathrm{hier}}(q,c_i),
\end{aligned}
\label{eq:fused_retrieval_score}
\end{equation}
where \(\lambda_1,\ldots,\lambda_4\) are fusion coefficients. This formulation prioritizes mandatory provisions over advisory or explanatory text when the candidates are otherwise similarly relevant.

A cross-encoder subsequently re-ranks the highest-scoring candidates:
\begin{equation}
S^{*}(q,c_i)
=
f_{\mathrm{CE}}(q,x_i).
\label{eq:cross_encoder_score}
\end{equation}
The resulting order determines the final textual evidence set, denoted by \(\mathcal{X}_q\).

\subsection{Query-Adaptive Visual Evidence Gating}
\label{subsec:visual_gating}

Attaching figures to answers that do not require visual evidence can reduce precision and increase unnecessary context. GEMS-RAG therefore employs a query-adaptive router that determines whether visual evidence should enter the generation stage:
\begin{equation}
\delta(q)\in\{0,1\},
\label{eq:visual_gate}
\end{equation}
where \(\delta(q)=1\) activates visual retrieval and \(\delta(q)=0\) suppresses all visual inputs.

Deterministic rules are evaluated first. An explicit figure or table identifier forces \(\delta(q)=1\), as does a visual expression such as ``show,'' ``appearance,'' or ``look like.'' In contrast, a definitional or permission-oriented question without visual cues forces \(\delta(q)=0\). Queries not resolved by the deterministic rules receive a visual-need score
\begin{equation}
s(q)
=
\sum_k
\beta_k
\ind{g_k(q)}
+
\beta_{\mathrm{kg}}
\frac{1}{
|\mathcal{C}_q|
}
\sum_{s\in\mathcal{C}_q}
\ind{
\operatorname{cites}(s)
},
\label{eq:visual_need_score}
\end{equation}
where \(g_k(q)\) represents lexical signals associated with device codes, visual nouns, and placement verbs. The set \(\mathcal{C}_q\) contains the sections associated with the highest-ranked textual chunks. The second term measures the proportion of those sections that cite at least one figure.

The router activates visual retrieval when
\begin{equation}
\delta(q)
=
\ind{
s(q)\geq\tau_{\mathrm{vis}}
},
\label{eq:gate_decision}
\end{equation}
where \(\tau_{\mathrm{vis}}\) is a decision threshold. The system records the rules and signals responsible for every routing decision. When \(\delta(q)=0\), neither figure crops nor page images are passed to the generator, and no images are included in the resulting answer.

\subsection{Visual Evidence Retrieval and Filtering}
\label{subsec:visual_retrieval}

When \(\delta(q)=1\), visual candidates are obtained through two complementary retrieval paths.

\paragraph{Graph-linked retrieval.}
The first path collects figures and tables cited by the highest-ranked textual chunks. These candidates are ordered according to the rank of their citing chunks.

\paragraph{Direct visual retrieval.}
The second path directly searches the visual crop index using late interaction. For a visual candidate \(F_j\), its relevance score is
\begin{equation}
S_{\mathrm{vis}}(q,F_j)
=
\sum_i
\max_t
\left\langle
\mathbf{q}_i,
\mathbf{E}_{j,t}
\right\rangle,
\label{eq:visual_late_interaction}
\end{equation}
where \(\mathbf{q}_i\) is the embedding of the \(i\)-th query token and \(\mathbf{E}_{j,t}\) is the \(t\)-th visual token representation of candidate \(F_j\).

The candidates returned by the two paths are merged and evaluated by a vision-language filtering model:
\begin{equation}
\mathcal{I}_q
=
f_{\mathrm{filter}}
\left(
q,
\{F_1,\ldots,F_m\}
\right),
\qquad
|\mathcal{I}_q|
\leq k_f,
\label{eq:visual_filter}
\end{equation}
where \(k_f\) is the maximum number of retained visual entities. The filter is instructed to retain only visual evidence that a reader must inspect to answer the query, with a conservative preference for excluding marginally relevant images. The filtering model may be smaller than the final answer-generation model. When a retained figure contains multiple sheets, all sheets associated with its canonical graph node are included.

\subsection{Multimodal Answer Generation}
\label{subsec:multimodal_generation}

The final evidence set is
\begin{equation}
\mathcal{R}(q)
=
\left\{
\mathcal{X}_q,\,
\delta(q)\mathcal{I}_q,\,
\delta(q)\mathcal{P}_q
\right\},
\label{eq:final_evidence_set}
\end{equation}
where \(\mathcal{X}_q\) denotes the textual evidence, \(\mathcal{I}_q\) denotes the filtered figure and table crops, and \(\mathcal{P}_q\) denotes the retrieved full-page renderings.

The multimodal generator receives the query, the textual evidence with section identifiers and rule types, and the gated visual evidence:
\begin{equation}
a
=
f_{\mathrm{VLM}}
\left(
q,
\mathcal{X}_q,
\delta(q)\mathcal{I}_q,
\delta(q)\mathcal{P}_q
\right).
\label{eq:multimodal_answer}
\end{equation}

Each textual chunk retains its section identifier and normative rule type, while each visual entity retains its canonical identifier and anchor section. After generation, every cited section, figure, and table is verified through an exact lookup in \(\mathcal{G}\). Citations that cannot be resolved to valid graph entities are rejected.

\subsection{Retrieval Procedure}
\label{subsec:retrieval_procedure}
For a query \(q\), GEMS-RAG extracts figure, table, section, and traffic-control-device identifiers and maps them to knowledge-graph nodes. It retrieves text using dense and learned sparse search, fuses rankings through reciprocal rank fusion, and scores candidates using graph proximity, rule-type weights, and hierarchy priors in Eq.~\eqref{eq:fused_retrieval_score}. The cross-encoder in Eq.~\eqref{eq:cross_encoder_score} then produces the textual evidence set \(\mathcal{X}_q\). The visual-evidence router determines \(\delta(q)\) and, when \(\delta(q)=1\), retrieves graph-linked and directly matched figures and tables for vision-language filtering. Finally, the framework assembles textual evidence, visual crops, captions, and page renderings into the generation context and produces an answer with graph-verified section, figure, table, and rule-type citations.

\section{Experimental Evaluation}
\begin{figure}[t]
    \centering
    \includegraphics[width=0.5\textwidth]{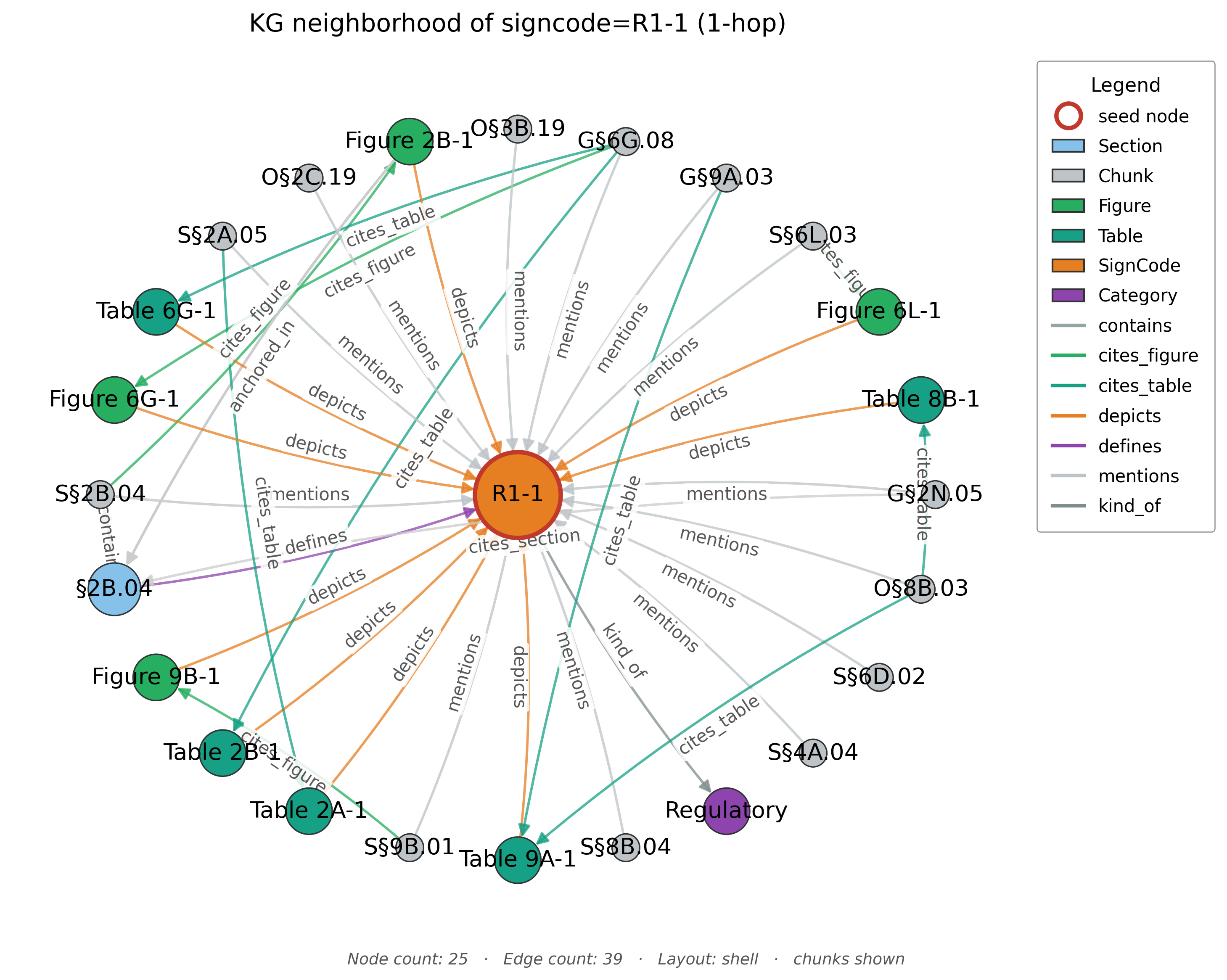}
    \caption{Knowledge graph for the R1-1 STOP sign example.}
    \label{fig:kg}
\end{figure}

This section evaluates GEMS-RAG on retrieval and question answering using the 11th Edition of the MUTCD. We assess its ability to retrieve sufficient textual and visual evidence, support grounded standards reasoning, and remain robust across evidence modalities and difficulty levels.
Table~\ref{tab:dataset_stats} summarizes the resulting knowledge base, while Figure~\ref{fig:kg} presents an exemplar subgraph centered on the STOP sign.

\begin{table}[t]
\centering
\caption{MUTCD dataset statistics.}
\label{tab:dataset_stats}
\begin{tabular}{lc}
\toprule
Component & Count \\
\midrule
Sections & 678 \\
Text Chunks & 5705 \\
Figures & 485 \\
Tables & 68 \\
Knowledge Graph Nodes & 8240 \\
Knowledge Graph Edges & 18245 \\
\bottomrule
\end{tabular}
\end{table}

\subsection{Evaluation Tasks and Results}
\begin{figure}[t]
    \centering
    \includegraphics[width=0.7\linewidth]
    {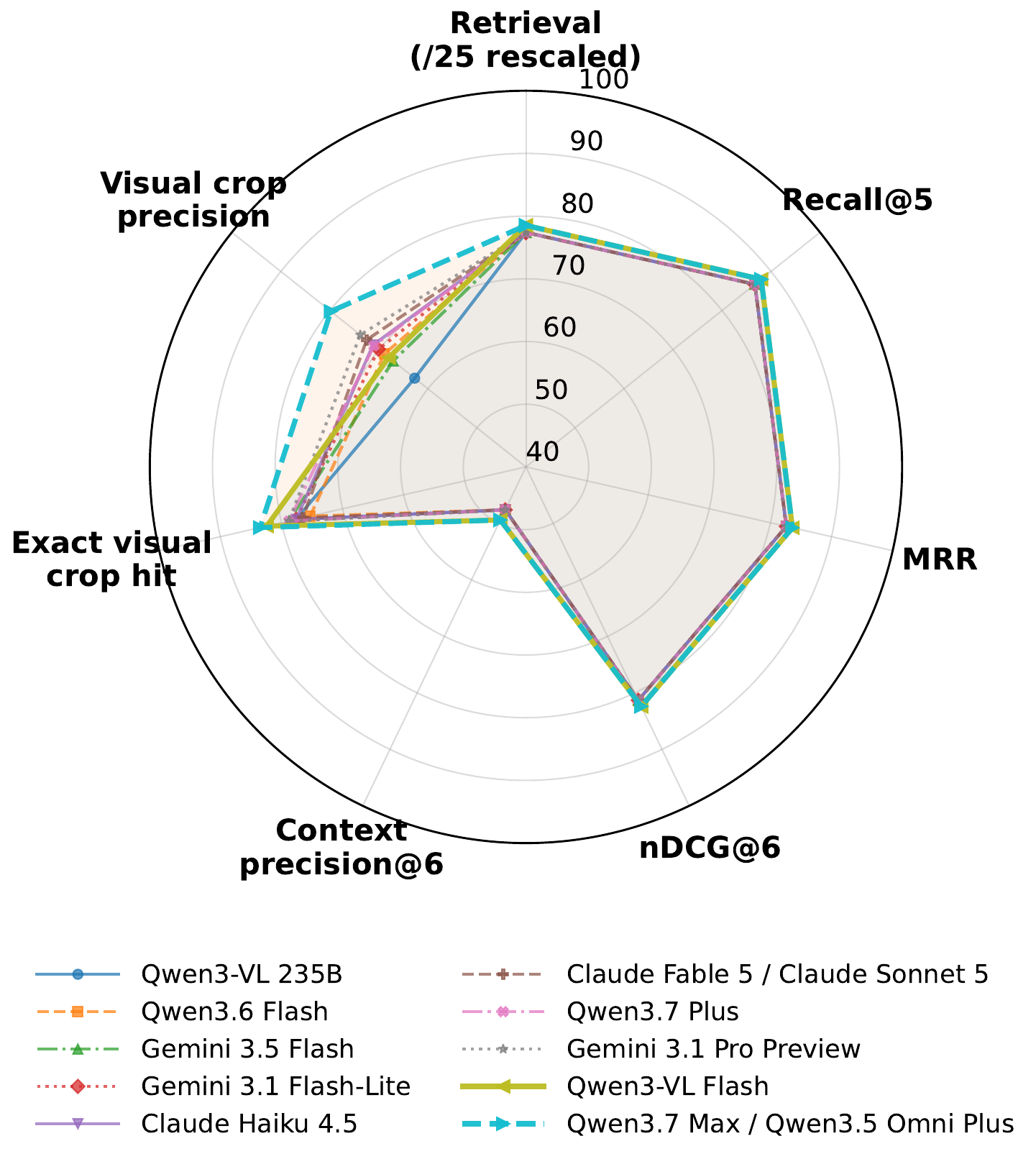}
    \caption{Retrieval and visual-evidence performance.}
    \label{fig:retrieval_performance}
\end{figure}

We constructed MUTCD-150 as a benchmark of 150 questions spanning textual, tabular, visual, and mixed-evidence reasoning. For each question, we manually locked the gold answer and the required MUTCD sections, source pages, tables, figures, and visual crops; 85 questions specifically required figure or table evidence. Using these fixed annotations and a common retrieval stack, we evaluated 12 vision-language models \citep{openai2026gpt56solpro,alibabacloud2026qwen36flash,google2026gemini35flash,google2026gemini31flashlite,google2026gemini31propreview,anthropic2025claudehaiku45,anthropic2026claudefable5,anthropic2026claudesonnet5,alibabacloud2026qwen37plus,alibabacloud2026qwen37max,alibabacloud2026qwen3vlflash,alibabacloud2026qwen35omniplus} on evidence retrieval, visual filtering, and question answering. Retrieval performance was consistent across models (19.35--19.63/25), with Recall@5 of 86.7\%--88.0\% and evidence sufficiency of 93.3\%. Larger differences emerged in model-controlled visual filtering, where exact visual-crop hit rates ranged from 75.3\% to 83.5\% and crop precision ranged from 62.7\% to 79.8\%, given in Figure~\ref{fig:retrieval_performance}. The shared retrieval stack produced stable evidence coverage across models, while the selected vision-language model primarily affected visual-crop selection. Context precision remained below 50\%, indicating that irrelevant evidence was still frequently passed to the answer-generation stage.

\begin{figure}[t]
    \centering
    \includegraphics[width=0.7\linewidth]
    {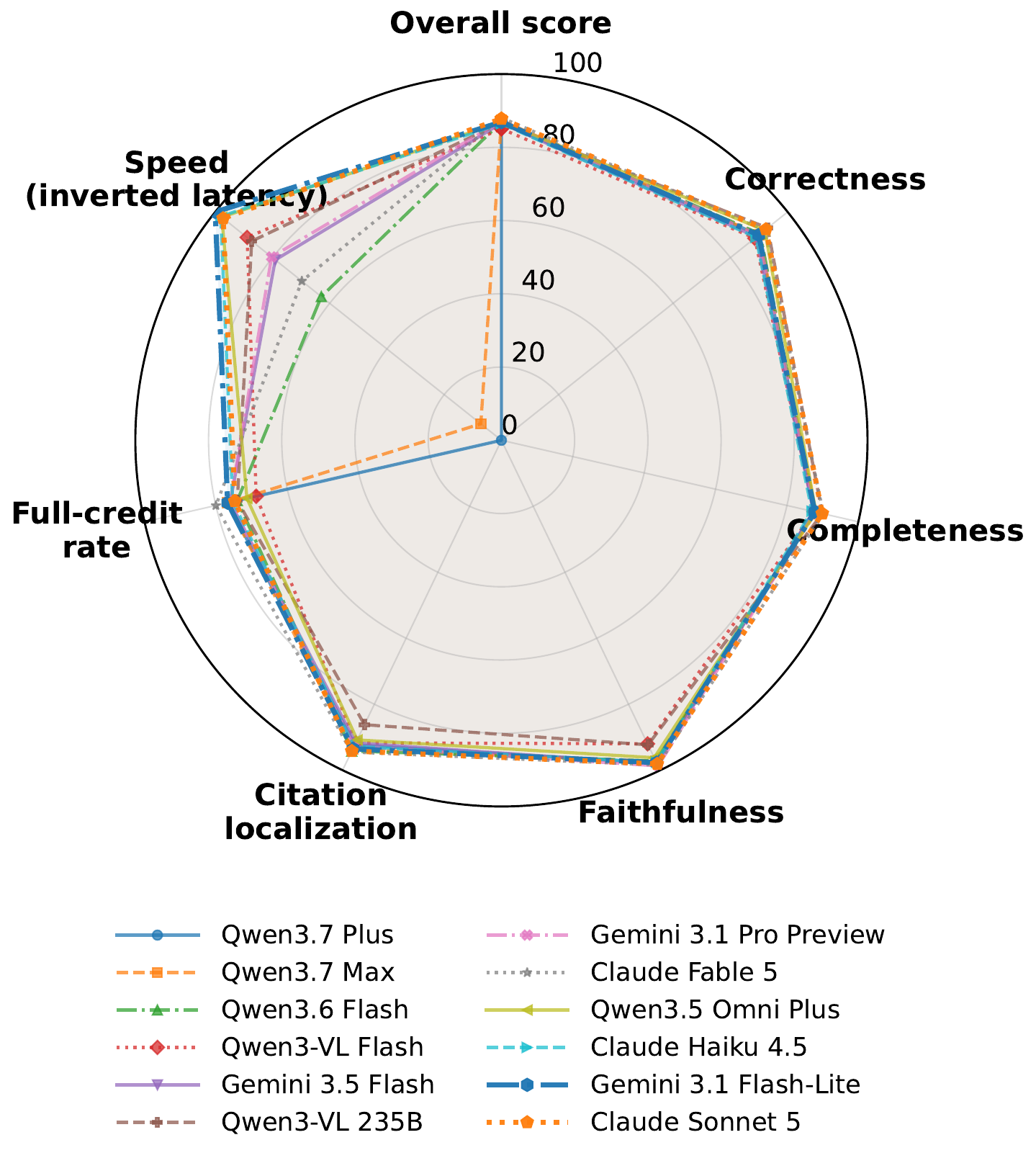}
    \caption{Question-answering performance.}
    \label{fig:qa_performance}
\end{figure}

For answer evaluation, GPT-5.6 \citep{openai2026gpt56solpro} served as a constrained, evidence-grounded LLM judge. Judgments with confidence above 50\% were accepted directly; lower-confidence cases were manually verified.
 It compared each model response with the locked gold answer and MUTCD evidence and applied a predefined rubric covering correctness, completeness, faithfulness, citation localization, and visual or table reasoning. Strict full credit required perfect correctness, completeness, faithfulness, and citation scores. As shown in Figure~\ref{fig:qa_performance}, overall normalized scores ranged from 85.08 to 88.08, with Claude Fable 5 achieving the highest provisional score and an 80.0\% full-credit rate. GPT-5.6 supported benchmark construction, evidence review, and rubric-based scoring, whereas retrieval metrics and final composite scores were calculated using fixed rule-based scripts.

\begin{figure}[t]
    \centering
    \includegraphics[width=\linewidth]{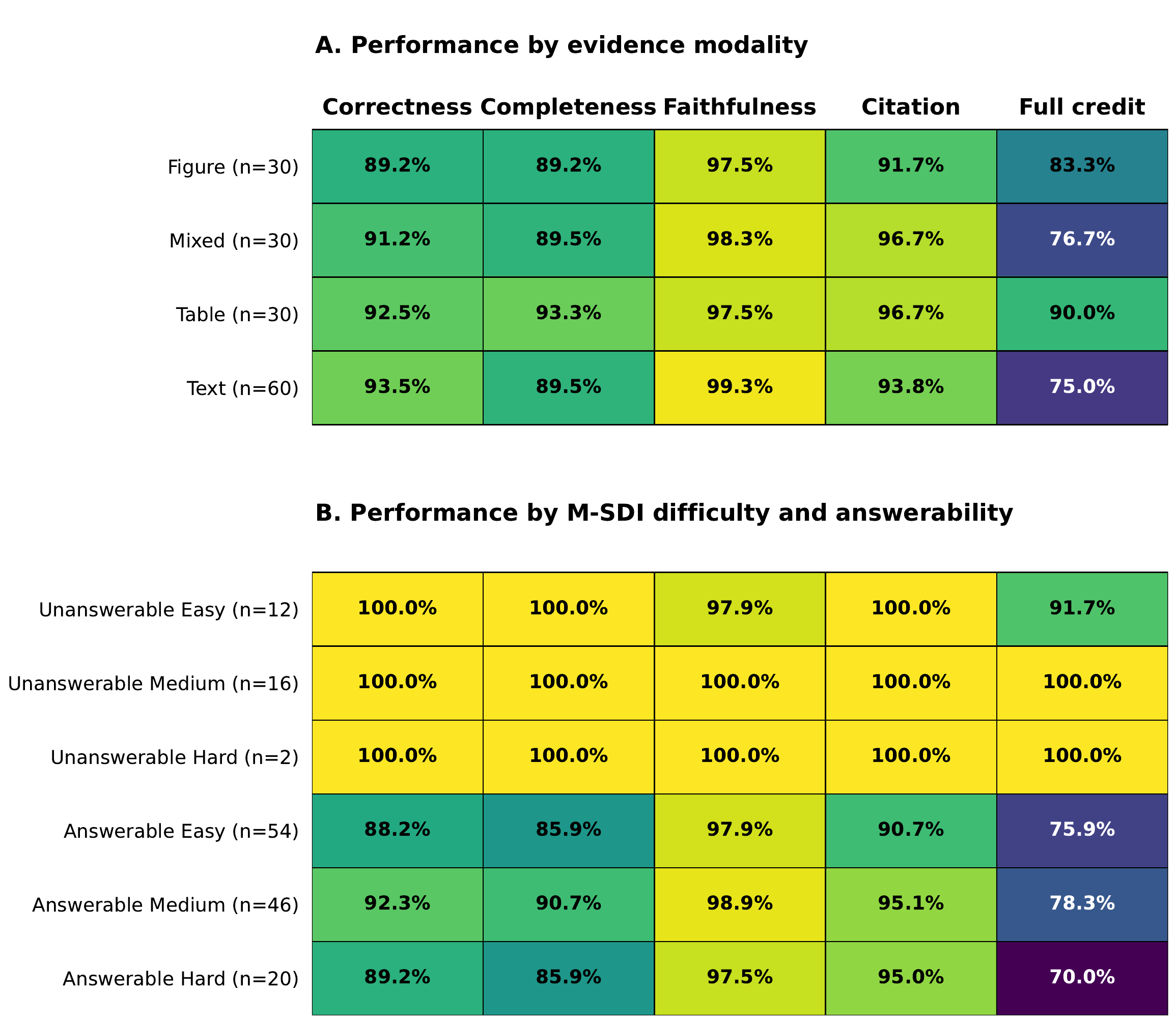}
    \caption{Claude Fable 5 performance by evidence modality and by M-SDI difficulty and answerability.}
    \label{fig:fable_breakdown}
\end{figure}

Based on the cross-model comparison, we selected Claude Fable 5 for detailed analysis because it achieved the highest provisional overall and strict full-credit scores while balancing correctness, faithfulness, citation localization, and visual reasoning. We then assessed its consistency across evidence modalities, answerability, and M-SDI difficulty levels, defined by localization, modality integration, reasoning, answer composition, and normative precision. As shown in Fig.~\ref{fig:fable_breakdown}, the model demonstrated strong performance across all evidence types. It achieved full credit on 90.0\% of tabular and 83.3\% of figure-based questions. Mixed-evidence questions yielded 91.2\% correctness, 89.5\% completeness, and a 76.7\% full-credit rate, while purely textual questions achieved high correctness (93.5\%) and faithfulness (99.3\%) alongside a 75.0\% full-credit rate. Notably, most point deductions stemmed from omitted definitions, exceptions, quantities, thresholds, or citations, rather than fundamentally incorrect conclusions.
Performance also remained stable across M-SDI difficulty levels. For answerable questions (those fully supported by MUTCD evidence), correctness reached 88.2\% on easy, 92.3\% on medium, and 89.2\% on hard queries. Furthermore, the model reliably handled unanswerable queries (those lacking sufficient, or in-scope support), maintaining near-perfect performance across all difficulty tiers.
\subsection{Performance Comparisons}
\begin{figure}[t]
    \centering
    \includegraphics[width=.7\linewidth]{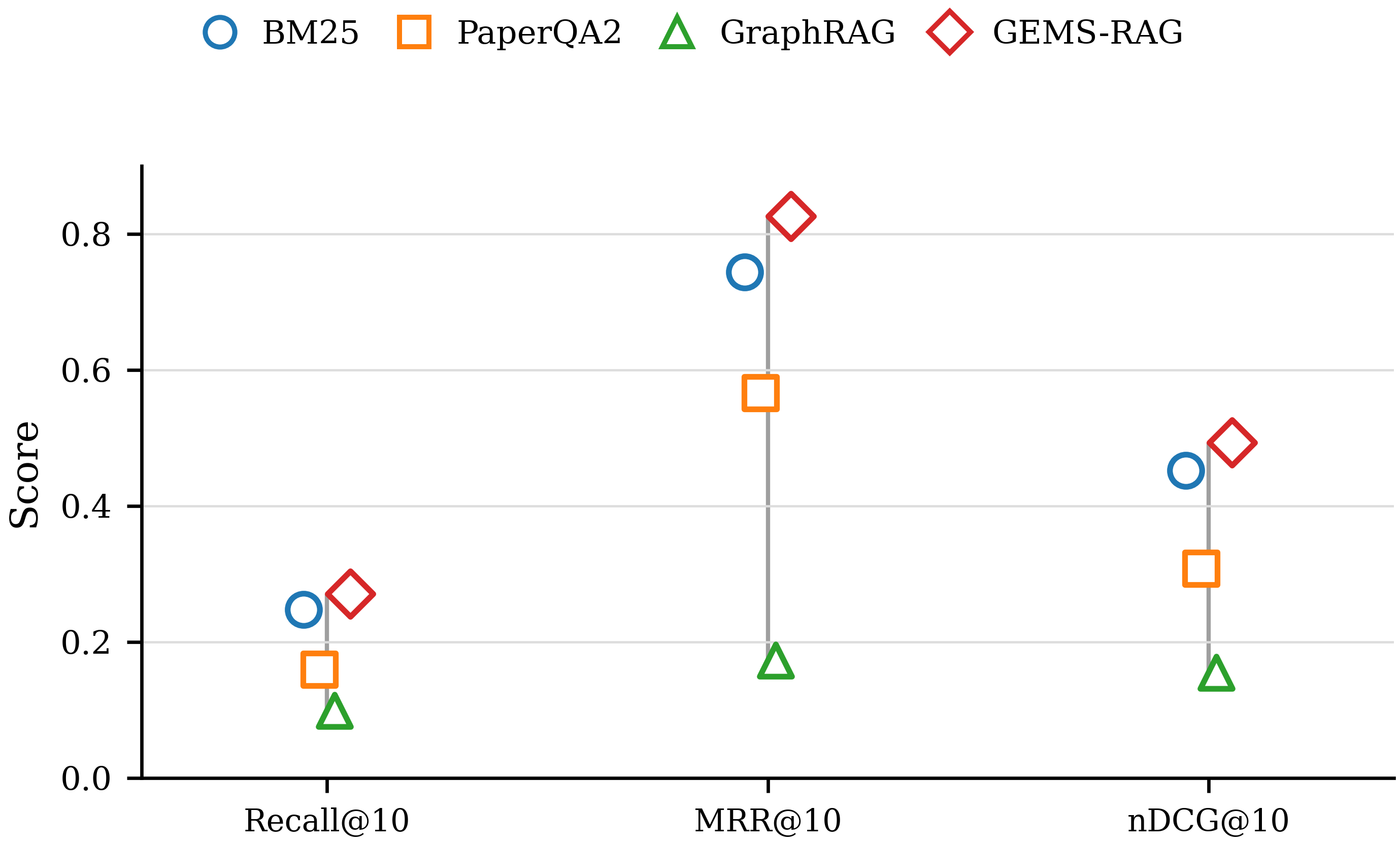}
    \caption{Retrieval performance comparison}
    \label{fig:retrieval_comp}
\end{figure}

\begin{figure}[t]
    \centering
    \includegraphics[width=1\linewidth]{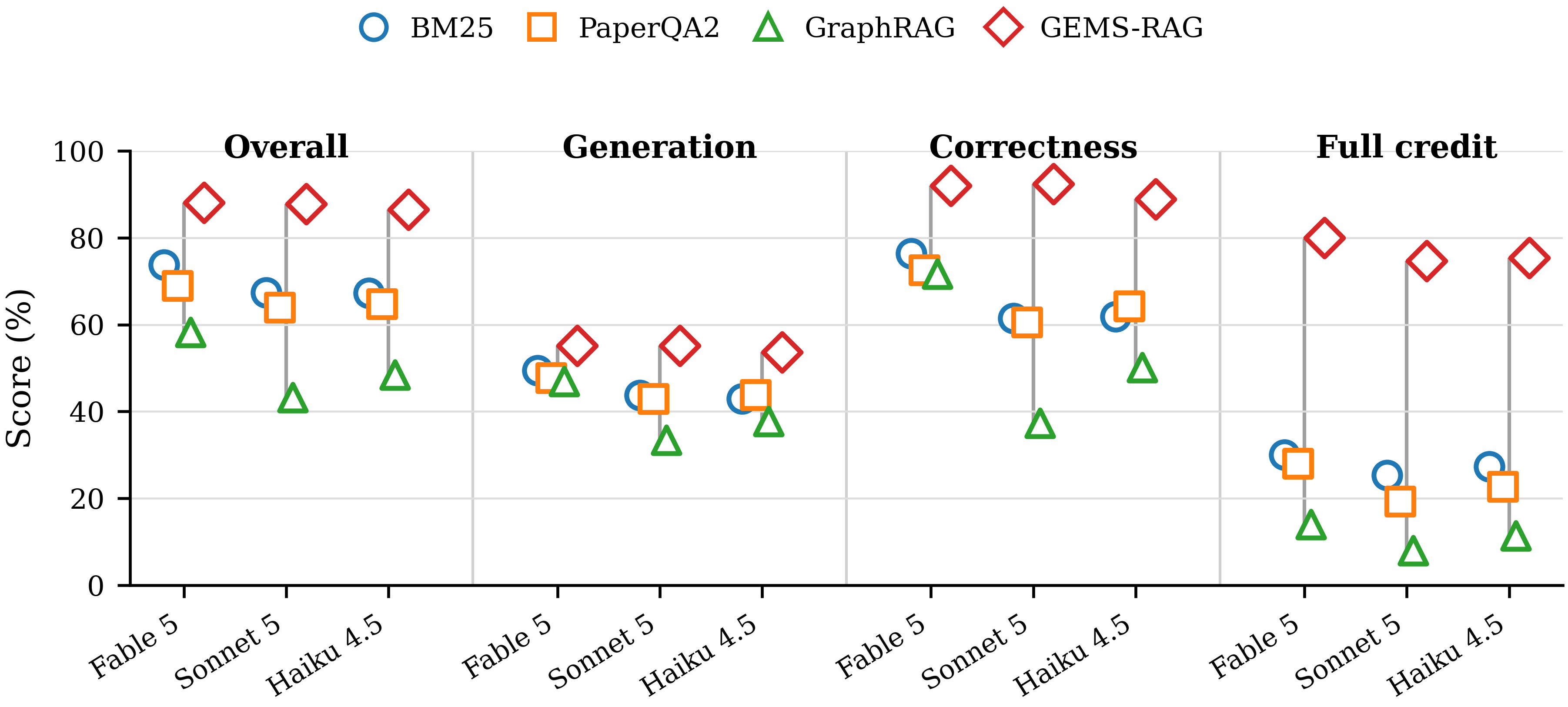}
    \caption{Question answering performance}
    \label{fig:comparison_data_plot}
\end{figure}

To evaluate GEMS-RAG, we benchmarked it against three established baselines (i.e., BM25 \citep{robertson2009probabilistic}, GraphRAG \citep{procko2024graph}, and PaperQA2 \citep{skarlinski2024language}) across three language models. We first assessed retrieval by comparing the top 10 figures and evidence items returned by each method against GPT-5.6-Pro generated gold annotations. As shown in Figure~\ref{fig:retrieval_comp}, GEMS-RAG achieves the highest Recall@10, MRR@10, and nDCG@10, whereas GraphRAG substantially underperforms BM25. As shown in Figure~\ref{fig:comparison_data_plot}, GEMS-RAG improves the overall score by more than 10\% across all models, with larger gains in correctness and full credit. Under the strict full-credit criterion, it exceeds the strongest baseline by approximately 48\%--50\% points, demonstrating its ability to generate correct, complete, and well-grounded answers.

\subsection{Ablation Study}
We assessed the contribution of three GEMS-RAG components through controlled ablations using the full Claude Fable 5 configuration as the reference.

As shown in Table~\ref{tab:ablation}, removing the VLM figure filter or hierarchy prior slightly increased retrieval scores but reduced generation, reliability, and overall performance. Removing the reranker caused the largest degradation, lowering the overall score by 4.65 points, which highlights its central role in selecting answer-relevant evidence.
%% AAAI kit: \setlength and \renewcommand are on the disallowed-commands list, so the
%% \tabcolsep and \arraystretch adjustments of the submission are removed; \resizebox is kept.
\begin{table}[t]
\centering
\caption{Ablation results. Best results are bolded and second-best results are underlined. Unit: percent.}
\label{tab:ablation}
\small
\resizebox{\columnwidth}{!}{%
\begin{tabular}{@{}lcccc@{}}
\toprule
Condition
& \makecell{Retr.\\(/25)}
& \makecell{Gen.\\(/60)}
& \makecell{Reliab.\\(/12)}
& \makecell{Overall\\(/100)} \\
\midrule

GEMS-RAG
& \underline{19.35}
& \textbf{55.18}
& \textbf{10.90}
& \textbf{88.08} \\

Without VLM filter
& \textbf{19.59} {\color{gray}(+0.24)}
& \underline{55.12} {\color{gray}(-0.06)}
& \underline{10.70} {\color{gray}(-0.20)}
& \underline{88.06} {\color{gray}(-0.02)} \\

Without hierarchy prior
& \textbf{19.59} {\color{gray}(+0.24)}
& 55.10 {\color{gray}(-0.08)}
& 10.54 {\color{gray}(-0.36)}
& 87.86 {\color{gray}(-0.22)} \\

Without reranker
& 17.78 {\color{gray}(-1.57)}
& 52.77 {\color{gray}(-2.41)}
& 10.38 {\color{gray}(-0.52)}
& 83.43 {\color{gray}(-4.65)} \\

\bottomrule
\end{tabular}%
}
\end{table}

\section{Conclusion}
This paper presented GEMS-RAG, a graph-grounded multimodal RAG framework for engineering standards. Unlike conventional pipelines that treat standards as independent text chunks, GEMS-RAG preserves document hierarchy, decomposes figures into fine-grained visual entities, and links provisions, figures, devices, rule types, and cross-references in a multimodal knowledge graph. By combining dense, sparse, and visual retrieval with graph expansion and rule-aware ranking, it enables standards-grounded question answering sensitive to visual specifications and regulatory structure.

Using the MUTCD as a representative engineering standards corpus, we constructed a multimodal knowledge base containing textual sections, figures, tables, visual entities, and graph relations. The experimental evaluation demonstrates that explicitly modeling multimodal and structural relationships can improve evidence retrieval, visual grounding, and downstream question answering compared with text-only, multimodal, and graph-based RAG baselines. The case study also illustrates that GEMS-RAG can retrieve not only the relevant textual provisions, but also associated visual entities, rule categories, and graph-connected context, leading to more complete and interpretable responses. Beyond the MUTCD, further work will apply GEMS-RAG to other engineering codes, design manuals and scientific standards.

%% ARXIV: acknowledgements were omitted from the anonymous submission. Fill in or delete.
%\section{Acknowledgments}
%[FUNDING AND ACKNOWLEDGEMENTS]

\bibliography{aaai2027}
\clearpage
\includepdf[pages=-]{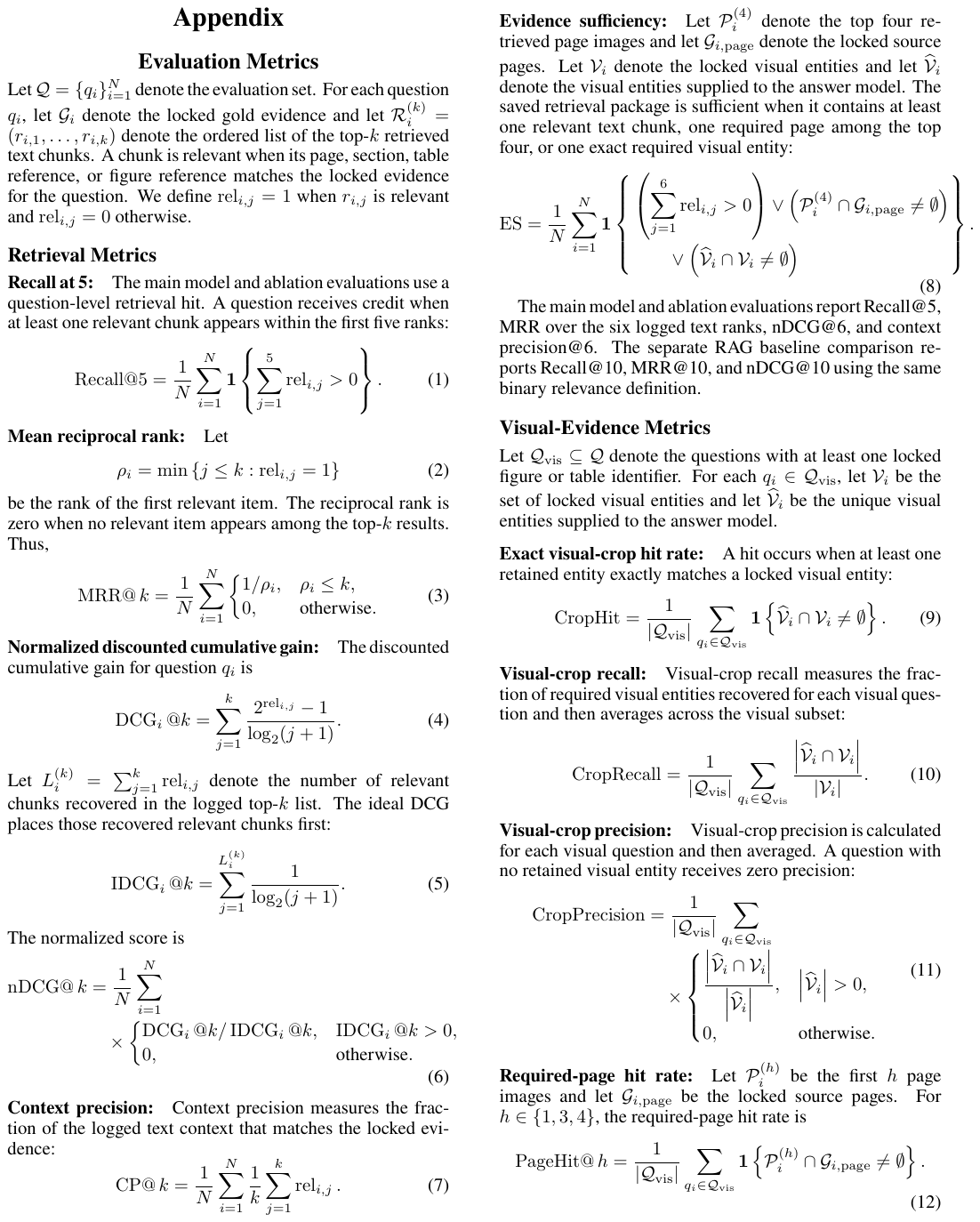}

\end{document}

\begin{center}
{\LARGE\bfseries Appendix}
\end{center}

% ----- MAIN DOCUMENT ---------------------------------------------
\appendix                      % switch to appendix mode
% -----------------------------------------------------------------
\section{Evaluation Metrics}
\label{app:evaluation_metrics}

Let $\mathcal{Q}=\{q_i\}_{i=1}^{N}$ denote the evaluation set. For each
question $q_i$, let $\mathcal{G}_i$ denote the locked gold evidence and
let $\mathcal{R}_i^{(k)}=(r_{i,1},\ldots,r_{i,k})$ denote the ordered
list of the top-$k$ retrieved text chunks. A chunk is relevant when its
page, section, table reference, or figure reference matches the locked
evidence for the question. We define $\operatorname{rel}_{i,j}=1$ when
$r_{i,j}$ is relevant and $\operatorname{rel}_{i,j}=0$ otherwise.

\subsection{Retrieval Metrics}

\paragraph{Recall at 5:}
The main model and ablation evaluations use a question-level retrieval
hit. A question receives credit when at least one relevant chunk appears
within the first five ranks:
\begin{equation}
\operatorname{Recall@5}
=
\frac{1}{N}
\sum_{i=1}^{N}
\mathbf{1}
\left\{
\sum_{j=1}^{5}
\operatorname{rel}_{i,j}>0
\right\}.
\label{eq:eval_recall}
\end{equation}

\paragraph{Mean reciprocal rank:}
Let
\begin{equation}
\rho_i
=
\min
\left\{
j\leq k:
\operatorname{rel}_{i,j}=1
\right\}
\label{eq:first_relevant_rank}
\end{equation}
be the rank of the first relevant item. The reciprocal rank is zero when
no relevant item appears among the top-$k$ results. Thus,
\begin{equation}
\operatorname{MRR@}k
=
\frac{1}{N}
\sum_{i=1}^{N}
\begin{cases}
1/\rho_i, & \rho_i\leq k,\\
0, & \text{otherwise}.
\end{cases}
\label{eq:eval_mrr}
\end{equation}

\paragraph{Normalized discounted cumulative gain:}
The discounted cumulative gain for question $q_i$ is
\begin{equation}
\operatorname{DCG}_i@k
=
\sum_{j=1}^{k}
\frac{
2^{\operatorname{rel}_{i,j}}-1
}{
\log_2(j+1)
}.
\label{eq:eval_dcg}
\end{equation}
Let $L_i^{(k)}=\sum_{j=1}^{k}\operatorname{rel}_{i,j}$ denote the number
of relevant chunks recovered in the logged top-$k$ list. The ideal DCG
places those recovered relevant chunks first:
\begin{equation}
\operatorname{IDCG}_i@k
=
\sum_{j=1}^{L_i^{(k)}}
\frac{1}{\log_2(j+1)}.
\label{eq:eval_idcg}
\end{equation}
The normalized score is
\begin{equation}
\begin{aligned}
\operatorname{nDCG@}k
&=
\frac{1}{N}
\sum_{i=1}^{N}
\\[-0.25em]
&\quad\times
\begin{cases}
\operatorname{DCG}_i@k/\operatorname{IDCG}_i@k,
& \operatorname{IDCG}_i@k>0,\\
0, & \text{otherwise}.
\end{cases}
\end{aligned}
\label{eq:eval_ndcg}
\end{equation}

\paragraph{Context precision:}
Context precision measures the fraction of the logged text context that
matches the locked evidence:
\begin{equation}
\operatorname{CP@}k
=
\frac{1}{N}
\sum_{i=1}^{N}
\frac{1}{k}
\sum_{j=1}^{k}
\operatorname{rel}_{i,j}.
\label{eq:context_precision}
\end{equation}

\paragraph{Evidence sufficiency:}
Let $\mathcal{P}_i^{(4)}$ denote the top four retrieved page images and
let $\mathcal{G}_{i,\mathrm{page}}$ denote the locked source pages. Let
$\mathcal{V}_i$ denote the locked visual entities and let
$\widehat{\mathcal{V}}_i$ denote the visual entities supplied to the
answer model. The saved retrieval package is sufficient when it contains
at least one relevant text chunk, one required page among the top four,
or one exact required visual entity:
\begin{equation}
\begin{aligned}
\operatorname{ES}
&=
\frac{1}{N}
\sum_{i=1}^{N}
\mathbf{1}
\left\{
\begin{aligned}
&\left(
\sum_{j=1}^{6}
\operatorname{rel}_{i,j}>0
\right)
\lor
\left(
\mathcal{P}_i^{(4)}
\cap
\mathcal{G}_{i,\mathrm{page}}
\neq
\emptyset
\right)
\\[-0.15em]
&\qquad\lor
\left(
\widehat{\mathcal{V}}_i
\cap
\mathcal{V}_i
\neq
\emptyset
\right)
\end{aligned}
\right\}.
\end{aligned}
\label{eq:evidence_sufficiency}
\end{equation}

The main model and ablation evaluations report Recall@5, MRR over the
six logged text ranks, nDCG@6, and context precision@6. The separate RAG
baseline comparison reports Recall@10, MRR@10, and nDCG@10 using the
same binary relevance definition.

\subsection{Visual-Evidence Metrics}

Let $\mathcal{Q}_{\mathrm{vis}}\subseteq\mathcal{Q}$ denote the
questions with at least one locked figure or table identifier. For each
$q_i\in\mathcal{Q}_{\mathrm{vis}}$, let $\mathcal{V}_i$ be the set of
locked visual entities and let $\widehat{\mathcal{V}}_i$ be the unique
visual entities supplied to the answer model.

\paragraph{Exact visual-crop hit rate:}
A hit occurs when at least one retained entity exactly matches a locked
visual entity:
\begin{equation}
\operatorname{CropHit}
=
\frac{1}{
|\mathcal{Q}_{\mathrm{vis}}|
}
\sum_{q_i\in\mathcal{Q}_{\mathrm{vis}}}
\mathbf{1}
\left\{
\widehat{\mathcal{V}}_i
\cap
\mathcal{V}_i
\neq
\emptyset
\right\}.
\label{eq:crop_hit}
\end{equation}

\paragraph{Visual-crop recall:}
Visual-crop recall measures the fraction of required visual entities
recovered for each visual question and then averages across the visual
subset:
\begin{equation}
\operatorname{CropRecall}
=
\frac{1}{
|\mathcal{Q}_{\mathrm{vis}}|
}
\sum_{q_i\in\mathcal{Q}_{\mathrm{vis}}}
\frac{
\left|
\widehat{\mathcal{V}}_i
\cap
\mathcal{V}_i
\right|
}{
\left|
\mathcal{V}_i
\right|
}.
\label{eq:crop_recall}
\end{equation}

\paragraph{Visual-crop precision:}
Visual-crop precision is calculated for each visual question and then
averaged. A question with no retained visual entity receives zero
precision:
\begin{equation}
\begin{aligned}
\operatorname{CropPrecision}
&=
\frac{1}{
|\mathcal{Q}_{\mathrm{vis}}|
}
\sum_{q_i\in\mathcal{Q}_{\mathrm{vis}}}
\\[-0.25em]
&\quad\times
\begin{cases}
\dfrac{
\left|
\widehat{\mathcal{V}}_i
\cap
\mathcal{V}_i
\right|
}{
\left|
\widehat{\mathcal{V}}_i
\right|
},
& \left|\widehat{\mathcal{V}}_i\right|>0,\\
0, & \text{otherwise}.
\end{cases}
\end{aligned}
\label{eq:crop_precision}
\end{equation}

\paragraph{Required-page hit rate:}
Let $\mathcal{P}_i^{(h)}$ be the first $h$ page images and let
$\mathcal{G}_{i,\mathrm{page}}$ be the locked source pages. For
$h\in\{1,3,4\}$, the required-page hit rate is
\begin{equation}
\operatorname{PageHit@}h
=
\frac{1}{
|\mathcal{Q}_{\mathrm{vis}}|
}
\sum_{q_i\in\mathcal{Q}_{\mathrm{vis}}}
\mathbf{1}
\left\{
\mathcal{P}_i^{(h)}
\cap
\mathcal{G}_{i,\mathrm{page}}
\neq
\emptyset
\right\}.
\label{eq:page_hit}
\end{equation}

\paragraph{Modality-routing accuracy:}
Let $y_i=1$ when locked visual evidence is required and $y_i=0$
otherwise. Given the router output $\delta(q_i)$, routing accuracy is
\begin{equation}
\operatorname{RoutingAcc}
=
\frac{1}{N}
\sum_{i=1}^{N}
\mathbf{1}
\left\{
\delta(q_i)=y_i
\right\}.
\label{eq:routing_accuracy}
\end{equation}

The visual metrics are reported as diagnostics. Exact visual hit and
visual precision do not receive separate points in the 25-point
retrieval score.

\subsection{Question-Answering Metrics}

For answer $a_i$, GPT-5.6 assigns normalized scores
$s_{i,d}\in[0,1]$ for correctness, completeness, faithfulness, and
citation localization. Let
\begin{equation}
\overline{M}_d
=
\frac{1}{N}
\sum_{i=1}^{N}
s_{i,d}
\label{eq:dimension_score}
\end{equation}
denote the mean score for dimension $d$. Reported percentages equal
$100\overline{M}_d$.

Visual reasoning is calculated as mean correctness over questions whose
primary modality is table, figure, or mixed:
\begin{equation}
\overline{M}_{\mathrm{vis}}
=
\frac{1}{|\mathcal{Q}_{\mathrm{nontext}}|}
\sum_{q_i\in\mathcal{Q}_{\mathrm{nontext}}}
s_{i,\mathrm{cor}}.
\label{eq:visual_reasoning}
\end{equation}
The relevance and concision term is held at
$\overline{M}_{\mathrm{rc}}=0.696$ for the completed model comparison.
The 60-point generation score is
\begin{equation}
\begin{split}
S_{\mathrm{gen}}
={}&
22\overline{M}_{\mathrm{cor}}
+
10\overline{M}_{\mathrm{com}}
+
12\overline{M}_{\mathrm{fai}}
+
6\overline{M}_{\mathrm{cit}}
\\
&+
7\overline{M}_{\mathrm{vis}}
+
3\overline{M}_{\mathrm{rc}}.
\end{split}
\label{eq:generation_score}
\end{equation}

\paragraph{Strict full credit:}
An answer receives full credit only when correctness, completeness,
faithfulness, and citation localization all equal 1:
\begin{equation}
\operatorname{FullCredit}
=
\frac{100}{N}
\sum_{i=1}^{N}
\prod_{
d\in
\{
\mathrm{cor},
\mathrm{com},
\mathrm{fai},
\mathrm{cit}
\}
}
\mathbf{1}
\left\{
s_{i,d}=1
\right\}.
\label{eq:full_credit}
\end{equation}

Scores by modality, answerability, or difficulty are calculated using
the same equations after restricting $\mathcal{Q}$ to the corresponding
question subset.

\subsection{Reliability Metrics}

Let $\mathcal{Q}_{\mathrm{unans}}$ denote the 30 questions whose locked
annotations mark them as unanswerable. Let $f_i$ equal 1 when answer
$a_i$ receives strict full credit and 0 otherwise. Unanswerable success
is
\begin{equation}
U
=
\frac{1}{|\mathcal{Q}_{\mathrm{unans}}|}
\sum_{q_i\in\mathcal{Q}_{\mathrm{unans}}}
f_i.
\label{eq:unanswerable_success}
\end{equation}

Let $c_i\in[0,1]$ denote the figure-router confidence recorded for
question $q_i$. The Brier score is
\begin{equation}
\operatorname{Brier}
=
\frac{1}{N}
\sum_{i=1}^{N}
\left(c_i-f_i\right)^2.
\label{eq:brier_score}
\end{equation}
For ten equal-width confidence bins $B_b$, expected calibration error is
\begin{equation}
\operatorname{ECE}_{10}
=
\sum_{b=1}^{10}
\frac{|B_b|}{N}
\left|
\frac{1}{|B_b|}
\sum_{i\in B_b}c_i
-
\frac{1}{|B_b|}
\sum_{i\in B_b}f_i
\right|.
\label{eq:ece_score}
\end{equation}
The calibration contribution and the measurable 12-point reliability
score are
\begin{equation}
S_{\mathrm{cal}}
=
6
\left[
1
-
\frac{
\operatorname{Brier}
+
\operatorname{ECE}_{10}
}{2}
\right]
\label{eq:calibration_score}
\end{equation}
and
\begin{equation}
S_{\mathrm{rel}}
=
6U
+
S_{\mathrm{cal}}.
\label{eq:reliability_score}
\end{equation}
The confidence value comes from the visual router rather than the final
answer model or GPT-5.6. The no-router ablation has no comparable
confidence value, so its Brier score, ECE, calibration points,
12-point reliability score, and 97-point full score are not defined.

\subsection{Composite Scores}

The 25-point retrieval score is
\begin{equation}
\begin{split}
S_{\mathrm{ret}}
={}&
10\operatorname{Recall@5}
+
5
\left(
\frac{
\operatorname{MRR}
+
\operatorname{nDCG@6}
}{2}
\right)
\\
&+
5\operatorname{CP@6}
+
3\operatorname{RoutingAcc}
+
2\operatorname{ES}.
\end{split}
\label{eq:retrieval_composite}
\end{equation}

Repeatability contributes up to 3 points in the intended 100-point
framework. That component remains pending. The measured score therefore
uses 97 available points:
\begin{equation}
S_{97}
=
S_{\mathrm{ret}}
+
S_{\mathrm{gen}}
+
S_{\mathrm{rel}},
\qquad
S_{97}\in[0,97].
\label{eq:measured_score}
\end{equation}
The reported provisional score normalizes the measured total to 100:
\begin{equation}
S_{\mathrm{provisional}}
=
100
\frac{S_{97}}{97}.
\label{eq:overall_score}
\end{equation}
The intended final score is
\begin{equation}
S_{\mathrm{final}}
=
S_{97}
+
S_{\mathrm{rep}},
\qquad
S_{\mathrm{rep}}\in[0,3],
\label{eq:final_score}
\end{equation}
where $S_{\mathrm{rep}}$ is the pending repeatability component.
Latency is reported separately through mean, median, and 90th-percentile
response time. It does not contribute to the composite score.

For the no-router ablation, the common score excludes calibration and
repeatability and uses a 91-point denominator:
\begin{equation}
S_{\mathrm{common}}
=
100
\frac{
S_{\mathrm{ret}}
+
S_{\mathrm{gen}}
+
6U
}{91}.
\label{eq:common_score}
\end{equation}

\subsection{MUTCD Structural Difficulty Index}

For each question $q_i$, the MUTCD Structural Difficulty Index
(M-SDI) evaluates five dimensions: evidence localization, modality
integration, reasoning operations, answer composition, and normative
precision. Let $h_{i,\ell}\in\{0,1,2\}$ denote the score for dimension
$\ell$. The total difficulty score is
\begin{equation}
\operatorname{M\text{-}SDI}_i
=
\sum_{\ell=1}^{5}h_{i,\ell},
\qquad
\operatorname{M\text{-}SDI}_i\in[0,10].
\label{eq:msdi}
\end{equation}
Questions are categorized as
\begin{equation}
\operatorname{Difficulty}(q_i)
=
\begin{cases}
\mathrm{Easy},   & 0\leq\operatorname{M\text{-}SDI}_i\leq3,\\
\mathrm{Medium}, & 4\leq\operatorname{M\text{-}SDI}_i\leq6,\\
\mathrm{Hard},   & 7\leq\operatorname{M\text{-}SDI}_i\leq10.
\end{cases}
\label{eq:msdi_categories}
\end{equation}

\subsection{Improvement over Baselines}

For GEMS-RAG score $S_{\mathrm{GEMS}}$ and baseline score $S_b$, the
relative improvement is
\begin{equation}
\Delta_{\mathrm{rel}}(b)
=
100
\frac{
S_{\mathrm{GEMS}}-S_b
}{
S_b
}.
\label{eq:relative_improvement}
\end{equation}
For percentage-valued metrics such as full-credit rate, the
percentage-point improvement over the strongest baseline is
\begin{equation}
\Delta_{\mathrm{pp}}
=
S_{\mathrm{GEMS}}
-
\max_{b\in\mathcal{B}}S_b.
\label{eq:percentage_point_gain}
\end{equation}

For the ablation study, we drew 800 paired bootstrap samples of the 150
question identifiers. Each sample used the same resampled questions for
the original RAG and the ablated condition. We recomputed the score
difference for each sample and reported the 2.5th and 97.5th percentiles.
These intervals measure uncertainty from the benchmark question sample.
They do not measure variation across repeated model calls.

\section{Reproducibility Assets}
\label{app:reproducibility_assets}

The benchmark assets were frozen before model execution. The runtime question
file contained only the question identifier and question text. A separate
locked gold file contained the gold answer, required sections, source pages,
figures, tables, answerability labels, and M-SDI annotations. The answering
models did not load the gold file. This separation prevented answer leakage and
kept the generation runs independent from the evaluation annotations.

The benchmark contained 150 questions, with 120 answerable and 30
unanswerable questions. The primary-modality distribution was 60 text, 30
table, 30 figure, and 30 mixed questions. The development split contained 30
questions and the test split contained 120 questions. The development split
supported prompt and retrieval configuration. The test split remained fixed for
final reporting.

We retained the source document, question-only benchmark, locked annotations,
model registry, runtime manifests, answer files, retrieval files, error logs,
item-level scores, ablation outputs, and per-response rationale records. Each
model run wrote separate JSONL files for answers and retrieval traces together
with a manifest that recorded the model identifier, run identifier, runner
version, and execution settings. These files provide the audit trail used to
reproduce the reported tables and figures.

Version control recorded the implementation state used for the experiments. The
updated ablation archives identify harness version \texttt{ablations\_v3} and
Git commit \texttt{ba26b07031398fb25efca}\allowbreak\texttt{8ba793a42c6ebb4ae03}.
The main benchmark records also identify runner version 1.1.2 and preserve the
exact provider model identifiers. Recording both the run manifest and source
revision distinguishes changes in model behavior from changes in the evaluation
code.

\section{Implementation Hyperparameters}
\label{app:hyperparameters}

The values in Tables~\ref{tab:rag_hyperparameters} and
\ref{tab:run_hyperparameters} come from the saved configuration, run manifests,
provider adapter, and ablation metadata. Parameters that the code did not set
explicitly are identified as provider defaults.

\begin{table*}[t]
\centering
\caption{Core indexing and retrieval hyperparameters.}
\label{tab:rag_hyperparameters}
\small
\begin{tabular}{p{0.27\textwidth} p{0.67\textwidth}}
\toprule
Setting & Value \\
\midrule
Text representation & \texttt{BAAI/bge-m3}, used for dense and learned sparse representations \\
Visual representation & \texttt{vidore/colqwen2-v0.1} \\
Cross-encoder reranker & \texttt{mixedbread-ai/mxbai-rerank-large-v2} \\
Page and figure rendering & 180 dpi for pages, 220 dpi for figure and table crops \\
Text retrieval depths & 30 dense, 30 sparse, 30 fused, 40 after graph expansion, 6 after reranking \\
Visual retrieval depths & 10 candidate entities, 6 direct visual candidates, at most 4 retained entities, 4 page images \\
Caption fallback & Disabled \\
Question router & Enabled, soft threshold 0.5, VLM tie break disabled \\
VLM figure filter & Enabled. For the final Claude and Gemini runs, the filter followed the selected answer model because the filter override was null \\
Fusion weights & Dense 1.00, sparse 0.60, hierarchy 0.20, graph proximity 0.40, rule type 0.30 \\
Rule-type multipliers & Standard 1.20, Guidance 1.00, Option 0.90, Support 0.70 \\
ColQwen2 image settings & Maximum 768 image patches, binary quantization enabled \\
Maximum text from each chunk & 1400 characters in the answer prompt \\
\bottomrule
\end{tabular}
\end{table*}

\begin{table*}[t]
\centering
\caption{Generation, benchmark-runner, and evaluation hyperparameters.}
\label{tab:run_hyperparameters}
\small
\begin{tabular}{p{0.27\textwidth} p{0.67\textwidth}}
\toprule
Setting & Value \\
\midrule
Detailed-analysis and ablation model & Claude Fable 5, alias \texttt{frontier\_claude} \\
Answer prompt & Few-shot, 3 examples \\
Visual-filter prompt & Zero-shot \\
Claude output-token cap & 16,000 tokens, including adaptive-thinking tokens and visible answer tokens \\
Truncation handling & One retry with a multiplier of 2.0, capped at 32,000 tokens for Claude \\
Temperature and top-p & Not set explicitly in the adapter, so the provider defaults applied \\
Benchmark runner & Version 1.1.2 \\
Random seed and replicates & Seed 20260710, 1 replicate per model configuration \\
Retry policy & Maximum 5 attempts, 2.0 seconds between requests, resume enabled \\
Ablation controls & Claude Fable 5 and the few-shot prompt remained fixed. Each run removed one component \\
Paired bootstrap & 800 paired resamples of the 150 question identifiers \\
\bottomrule
\end{tabular}
\end{table*}

\subsection{Retry, Resume, and Failure Handling}

The runner wrote each completed record to disk before moving to the next
question. Resume mode skipped valid records that already existed and continued
from missing or failed question identifiers. The runner allowed at most five
provider attempts and placed a 2.0-second interval between requests. For Claude,
the provider adapter used a 16,000-token output cap and performed one
truncation-specific retry with twice the limit, capped at 32,000 tokens.

We reran records only for operational failures, missing records, corrupted
outputs, provider errors, or materially incomplete retrieval logging. We did
not rerun an answer because it was factually incorrect or received a low score.
When a retry succeeded, the evaluator retained the first valid response for the
question and preserved the original failure record for audit. This rule avoided
selecting a more favorable answer after observing model quality.

\subsection{Role of GPT-5.6 and the Evaluated Language Models}

The authors used GPT-5.6
to review benchmark-runner code, provider adapters, runtime validation
checks, scoring scripts, plotting code, and technical documentation. The authors
executed the code, inspected saved records, selected the final settings, and
resolved failures before accepting an output. Model-generated code or prose did
not enter the reported results without execution and record-level verification.

During evaluation, GPT-5.6 supported benchmark construction, evidence review,
and application of the locked answer-quality rubric. It did not generate the
retrieval ranks, latency values, calibration statistics, or composite scores.
Fixed scripts calculated those quantities from the saved records. The evaluated
vision-language models received only the user question and retrieved evidence.
They did not receive the locked gold answer file.

The cross-model study used twelve vision-language configurations under the same
benchmark and scoring framework. The selected model also controlled visual
filtering when the filter-model override was null. Claude Fable 5 produced the
answers for the detailed modality, difficulty, and ablation analyses after it
achieved the highest provisional score in the cross-model comparison. The
ablation study kept Claude Fable 5 and the few-shot prompt fixed while removing
one retrieval component at a time.

\section{Answer-Rationale Audit and Evaluator Confidence}
\label{app:evaluator_confidence}

We created one grading rationale for each of the 900 ablation responses and
each of the 150 original-RAG baseline responses. Each record retained the four
official item scores, the strict full-credit flag, the model answer, the locked
gold answer, the retrieved evidence metrics, and the rationale provenance. The
rationale package distinguishes an explicit adjudication note from a structured
reconstruction of an existing grade.

\subsection{Programmatic Consistency Checks}

Table~\ref{tab:rationale_checks} summarizes checks performed over all 1,050
records. These checks validate the internal representation of the official
grades. They do not establish that every underlying item grade is correct
against the MUTCD.

\begin{table*}[t]
\centering
\caption{Programmatic checks of the per-response rationale package.}
\label{tab:rationale_checks}
\footnotesize
\begin{tabular}{p{0.57\textwidth} p{0.37\textwidth}}
\toprule
Check & Verified result \\
\midrule
Strict full-credit flag compared with all four dimensions equal to 1.00 & 0 mismatches in 1,050 records \\
Descriptive mean compared with the mean of correctness, completeness, faithfulness, and citation & 0 mismatches in 1,050 records \\
Per-condition correctness, completeness, faithfulness, citation, and full-credit aggregates compared with the version 4 report & Exact match for all seven conditions \\
Strict full-credit counts & 120, 114, 112, 114, 112, 109, and 103 for the baseline and A1 through A6 \\
Rationale provenance & 895 structured reconstructions, 48 explicit adjudication notes, and 107 reused baseline notes \\
Reused-note criterion & 0 violations. Each reused note had the same question identifier and identical four official scores as the baseline \\
Explicit-note allocation & 30 original-RAG partial-credit records and 18 repaired A1 records \\
Repaired A1 records & Exactly T001 through T018 \\
\bottomrule
\end{tabular}
\end{table*}

The token-overlap field is diagnostic rather than authoritative. It listed a
missing or underdeveloped element in 216 full-credit records, primarily when an
unanswerable response used a refusal phrasing that did not lexically match the
gold wording. It listed no missing element in 173 partial-credit records, where
the deduction instead came from an adjudication note, faithfulness, or citation
localization. The official four scores and the adjudication notes remain the
source of truth.

A direct recheck found 20 records with citation score 1.00 and an empty
regex-detected citation field. All 20 were unanswerable responses involving a
false premise, unsupported precision, or out-of-scope request. The accepted
citation-localization score reflected correct handling of the unsupported
request, while the citation detector searched only for explicit section,
figure, table, page, or legal-reference patterns. We therefore treat these
records as a detector limitation rather than a contradiction in the official
grade.

\subsection{Evaluator Confidence}

GPT-5.6 gave a strict self-assessment of 82\% confidence in the complete
ablation evaluation. This value reflected high confidence in deterministic
runtime and retrieval calculations and lower confidence in rubric-based item
judgments. The value is a judgmental assessment, not a model probability and
not a sampling-based confidence interval.

The authors then supplied the 1,050 rationale records to Claude Fable 5 for a
secondary model audit. Fable 5 checked score arithmetic, aggregate agreement,
provenance rules, repaired-record allocation, and a small set of MUTCD content
examples. It assigned 85\% confidence with a reported evaluator-confidence
interval of [78\%, 90\%]. Fable 5 placed the mechanical consistency of the
package near the top of its scale because the score and aggregate checks
matched. It reduced its overall assessment because most rationales reconstruct
or restate existing grades and therefore do not independently validate the
substantive MUTCD judgment.

The reported evaluator-confidence summary is
\begin{equation}
\widehat{C}_{\mathrm{Fable}}=0.85,
\qquad
I_{\mathrm{Fable}}=[0.78,0.90].
\label{eq:fable_confidence}
\end{equation}
This interval is a judgmental interval supplied by a secondary language-model
auditor. It is not a frequentist confidence interval, a Bayesian credible
interval, or the paired bootstrap interval used for ablation score differences.
Claude Fable 5 also generated the evaluated answers, so this secondary audit
does not replace an independent MUTCD-informed human annotator.

The rationale package assigns fixed confidence values by provenance tier:
0.90 for explicit adjudication notes, 0.88 for reconstructed full-credit
records, 0.82 for the 18 repaired A1 responses reviewed by GPT-5.6, and 0.76
for reconstructed partial-credit rationales. These are design constants used
to describe confidence in the explanation. They are not empirical
probabilities. A second MUTCD-informed annotator should review a stratified
sample, report agreement, and adjudicate disagreements before the item-level
judgments are treated as independently validated.

\section{Sample MUTCD-150 Questions and Locked Gold Answers}
\label{app:qa_samples}

This section provides 10 representative examples for each reporting
category: answerable, unanswerable, easy, medium, and hard. The five
displayed sets do not repeat question identifiers. Answerability and
M-SDI difficulty remain independent benchmark labels, so each entry
retains both labels. We selected examples to cover text, table, figure,
and mixed evidence where the category permits. The question wording,
locked gold answers, evidence identifiers, answerability labels, and
M-SDI scores below are copied from the locked
gold file. The answering
models did not receive this file during execution.

\subsection{Answerable Questions}

\begingroup
\setlength{\fboxsep}{5pt}
\noindent\fcolorbox{blue!60!black}{blue!60!black}{%
\parbox{\dimexpr\linewidth-2\fboxsep-2\fboxrule\relax}{%
\color{white}\centering\bfseries\strut Answerable Questions: 10 Locked Examples%
}}%
\par\vspace{-\fboxrule}
\noindent\fcolorbox{blue!60!black}{blue!7!white}{%
\parbox{\dimexpr\linewidth-2\fboxsep-2\fboxrule\relax}{%
\small
\noindent\textbf{1. T002}\hfill\textit{Text, Easy, Answerable, M-SDI 1/10}\\[2pt]
\textbf{Question}: Does the applicability of the MUTCD to a facility open to public travel depend on whether the facility is publicly or privately owned or on the source of its funding?\\[2pt]
\textbf{Locked Gold Answer}: No. Applicability is independent of public or private ownership or jurisdiction and independent of whether funding is Federal, State, local, or private.\\[2pt]
\textbf{Locked Evidence}: Section 1A.01, PDF page 42.
}}%
\par\vspace{-\fboxrule}
\noindent\fcolorbox{blue!60!black}{blue!7!white}{%
\parbox{\dimexpr\linewidth-2\fboxsep-2\fboxrule\relax}{%
\small
\noindent\textbf{2. T005}\hfill\textit{Text, Easy, Answerable, M-SDI 3/10}\\[2pt]
\textbf{Question}: How do the MUTCD headings Standard, Guidance, Option, and Support differ, including the verbs typically associated with them?\\[2pt]
\textbf{Locked Gold Answer}: A Standard is mandatory or prohibitive and typically uses 'shall.' Guidance is recommended practice with justified deviations allowed and typically uses 'should.' An Option is permissive and typically uses 'may.' Support is informational and conveys no mandate, recommendation, permission, prohibition, or enforceable condition.\\[2pt]
\textbf{Locked Evidence}: Section 1C.01, PDF page 54.
}}%
\par\vspace{-\fboxrule}
\noindent\fcolorbox{blue!60!black}{blue!7!white}{%
\parbox{\dimexpr\linewidth-2\fboxsep-2\fboxrule\relax}{%
\small
\noindent\textbf{3. T025}\hfill\textit{Text, Medium, Answerable, M-SDI 5/10}\\[2pt]
\textbf{Question}: What must a Speed Limit sign display, in what increments, and may it be used to show an advisory speed for a roadway condition?\\[2pt]
\textbf{Locked Gold Answer}: It must display the speed limit established by law, ordinance, regulation, or an authorized agency based on an engineering study, in multiples of 5 mph. It should not be used for an advisory speed; an advisory speed plaque below an appropriate warning sign should be used instead.\\[2pt]
\textbf{Locked Evidence}: Section 2B.21, PDF pages 126, 127.
}}%
\par\vspace{-\fboxrule}
\noindent\fcolorbox{blue!60!black}{blue!7!white}{%
\parbox{\dimexpr\linewidth-2\fboxsep-2\fboxrule\relax}{%
\small
\noindent\textbf{4. TB001}\hfill\textit{Table, Easy, Answerable, M-SDI 3/10}\\[2pt]
\textbf{Question}: According to Table 1B-1, what is the compliance date for implementing and continuing to use a method designed to maintain longitudinal pavement-marking retroreflectivity?\\[2pt]
\textbf{Locked Gold Answer}: September 6, 2026.\\[2pt]
\textbf{Locked Evidence}: Section 1B.03, Table 1B-1, PDF page 46.
}}%
\par\vspace{-\fboxrule}
\noindent\fcolorbox{blue!60!black}{blue!7!white}{%
\parbox{\dimexpr\linewidth-2\fboxsep-2\fboxrule\relax}{%
\small
\noindent\textbf{5. TB002}\hfill\textit{Table, Medium, Answerable, M-SDI 4/10}\\[2pt]
\textbf{Question}: In Table 1D-1, what standard abbreviations are permitted for 'Pedestrian(s)'?\\[2pt]
\textbf{Locked Gold Answer}: PED or PEDS.\\[2pt]
\textbf{Locked Evidence}: Section 1D.08, Table 1D-1, PDF page 76.
}}%
\par\vspace{-\fboxrule}
\noindent\fcolorbox{blue!60!black}{blue!7!white}{%
\parbox{\dimexpr\linewidth-2\fboxsep-2\fboxrule\relax}{%
\small
\noindent\textbf{6. TB024}\hfill\textit{Table, Medium, Answerable, M-SDI 5/10}\\[2pt]
\textbf{Question}: What minimum off-roadway sizes does Table 9A-1 give for a STOP sign, a YIELD sign, and a Crossbuck sign on a bicycle facility?\\[2pt]
\textbf{Locked Gold Answer}: STOP: 18 x 18 inches; YIELD: 18 x 18 x 18 inches; Crossbuck: 24 x 4.5 inches.\\[2pt]
\textbf{Locked Evidence}: Section 9A.03, Table 9A-1, PDF page 1090.
}}%
\par\vspace{-\fboxrule}
\noindent\fcolorbox{blue!60!black}{blue!7!white}{%
\parbox{\dimexpr\linewidth-2\fboxsep-2\fboxrule\relax}{%
\small
\noindent\textbf{7. F008}\hfill\textit{Figure, Medium, Answerable, M-SDI 5/10}\\[2pt]
\textbf{Question}: From Figure 2B-1, identify the shapes of the STOP and YIELD signs and the color scheme of the ALL-WAY plaque.\\[2pt]
\textbf{Locked Gold Answer}: STOP is an octagon; YIELD is a downward-pointing equilateral triangle; the ALL-WAY plaque has white legend and border on a red background.\\[2pt]
\textbf{Locked Evidence}: Sections 2B.04, 2B.05, Figure 2B-1, PDF page 115.
}}%
\par\vspace{-\fboxrule}
\noindent\fcolorbox{blue!60!black}{blue!7!white}{%
\parbox{\dimexpr\linewidth-2\fboxsep-2\fboxrule\relax}{%
\small
\noindent\textbf{8. F012}\hfill\textit{Figure, Medium, Answerable, M-SDI 5/10}\\[2pt]
\textbf{Question}: What four crosswalk marking patterns are illustrated in Figure 3C-1?\\[2pt]
\textbf{Locked Gold Answer}: Transverse, longitudinal bar, ladder, and bar pair.\\[2pt]
\textbf{Locked Evidence}: Section 3C.03, Figure 3C-1, PDF page 631.
}}%
\par\vspace{-\fboxrule}
\noindent\fcolorbox{blue!60!black}{blue!7!white}{%
\parbox{\dimexpr\linewidth-2\fboxsep-2\fboxrule\relax}{%
\small
\noindent\textbf{9. M001}\hfill\textit{Mixed, Hard, Answerable, M-SDI 7/10}\\[2pt]
\textbf{Question}: A provision appears under a Guidance heading and uses the word 'should.' Is it mandatory, and when may a deviation be appropriate?\\[2pt]
\textbf{Locked Gold Answer}: It is recommended practice, not a mandatory Standard. A deviation may be appropriate when engineering judgment or an engineering study indicates that the deviation is suitable for the specific situation.\\[2pt]
\textbf{Locked Evidence}: Sections 1A.04, 1C.01, PDF pages 43, 54.
}}%
\par\vspace{-\fboxrule}
\noindent\fcolorbox{blue!60!black}{blue!7!white}{%
\parbox{\dimexpr\linewidth-2\fboxsep-2\fboxrule\relax}{%
\small
\noindent\textbf{10. M019}\hfill\textit{Mixed, Hard, Answerable, M-SDI 9/10}\\[2pt]
\textbf{Question}: For a 12-foot lateral offset and a 35-mph speed, calculate L using Table 6B-4. Then state the minimum merging, shifting, and shoulder taper lengths from Table 6B-3.\\[2pt]
\textbf{Locked Gold Answer}: For 40 mph or less, L = W*S$^2$/60 = 12*35$^2$/60 = 245 feet. Minimum merging taper: 245 feet; shifting taper: 122.5 feet; shoulder taper: about 80.9 feet (0.33L, commonly rounded appropriately in design).\\[2pt]
\textbf{Locked Evidence}: Section 6B.08, Tables 6B-3, 6B-4, PDF page 816.
}}%
\par
\endgroup

\subsection{Unanswerable Questions}

\begingroup
\setlength{\fboxsep}{5pt}
\noindent\fcolorbox{gray!65!black}{gray!65!black}{%
\parbox{\dimexpr\linewidth-2\fboxsep-2\fboxrule\relax}{%
\color{white}\centering\bfseries\strut Unanswerable Questions: 10 Locked Examples%
}}%
\par\vspace{-\fboxrule}
\noindent\fcolorbox{gray!65!black}{gray!8!white}{%
\parbox{\dimexpr\linewidth-2\fboxsep-2\fboxrule\relax}{%
\small
\noindent\textbf{1. T049}\hfill\textit{Text, Medium, Unanswerable, M-SDI 5/10, Unsupported Precision}\\[2pt]
\textbf{Question}: According to the MUTCD, what exact percentage reduction in crashes is guaranteed by installing a STOP sign or all-way stop control?\\[2pt]
\textbf{Locked Gold Answer}: The MUTCD does not specify this information. A correct response should state that the requested value or requirement cannot be determined from the manual rather than inventing an answer.\\[2pt]
\textbf{Locked Evidence}: Sections 2B.04, 2B.12-2B.17, PDF pages 115, 120.
}}%
\par\vspace{-\fboxrule}
\noindent\fcolorbox{gray!65!black}{gray!8!white}{%
\parbox{\dimexpr\linewidth-2\fboxsep-2\fboxrule\relax}{%
\small
\noindent\textbf{2. T052}\hfill\textit{Text, Medium, Unanswerable, M-SDI 4/10, False Presupposition}\\[2pt]
\textbf{Question}: What single mandatory signal-cycle length must be used at every urban signalized intersection in the United States?\\[2pt]
\textbf{Locked Gold Answer}: The MUTCD does not specify this information. A correct response should state that the requested value or requirement cannot be determined from the manual rather than inventing an answer.\\[2pt]
\textbf{Locked Evidence}: Sections 4A.01-4A.06, 4B.01-4B.05, PDF pages 684, 691.
}}%
\par\vspace{-\fboxrule}
\noindent\fcolorbox{gray!65!black}{gray!8!white}{%
\parbox{\dimexpr\linewidth-2\fboxsep-2\fboxrule\relax}{%
\small
\noindent\textbf{3. T054}\hfill\textit{Text, Medium, Unanswerable, M-SDI 5/10, Out Of Scope}\\[2pt]
\textbf{Question}: How many licensed traffic engineers must a road agency employ per 100 centerline miles to comply with the MUTCD?\\[2pt]
\textbf{Locked Gold Answer}: The MUTCD does not specify this information. A correct response should state that the requested value or requirement cannot be determined from the manual rather than inventing an answer.\\[2pt]
\textbf{Locked Evidence}: Sections 1A.04, 1D.03, PDF pages 43, 73.
}}%
\par\vspace{-\fboxrule}
\noindent\fcolorbox{gray!65!black}{gray!8!white}{%
\parbox{\dimexpr\linewidth-2\fboxsep-2\fboxrule\relax}{%
\small
\noindent\textbf{4. T059}\hfill\textit{Text, Easy, Unanswerable, M-SDI 3/10, Contradicted Premise}\\[2pt]
\textbf{Question}: What minimum miles-per-gallon rating must a vehicle achieve to qualify as an inherently low-emission vehicle under the MUTCD?\\[2pt]
\textbf{Locked Gold Answer}: The MUTCD does not specify this information. A correct response should state that the requested value or requirement cannot be determined from the manual rather than inventing an answer.\\[2pt]
\textbf{Locked Evidence}: Section 1C.02, PDF pages 60, 62.
}}%
\par\vspace{-\fboxrule}
\noindent\fcolorbox{gray!65!black}{gray!8!white}{%
\parbox{\dimexpr\linewidth-2\fboxsep-2\fboxrule\relax}{%
\small
\noindent\textbf{5. TB027}\hfill\textit{Table, Easy, Unanswerable, M-SDI 3/10, Unsupported Precision}\\[2pt]
\textbf{Question}: What RGB values are listed in Table 2A-2 for fluorescent yellow-green?\\[2pt]
\textbf{Locked Gold Answer}: The MUTCD does not specify this information. A correct response should state that the requested value or requirement cannot be determined from the manual rather than inventing an answer.\\[2pt]
\textbf{Locked Evidence}: Section 2A.06, Table 2A-2, PDF page 85.
}}%
\par\vspace{-\fboxrule}
\noindent\fcolorbox{gray!65!black}{gray!8!white}{%
\parbox{\dimexpr\linewidth-2\fboxsep-2\fboxrule\relax}{%
\small
\noindent\textbf{6. TB029}\hfill\textit{Table, Medium, Unanswerable, M-SDI 6/10, Out Of Scope}\\[2pt]
\textbf{Question}: Which table lists State-by-State monetary fines for violating a No Turn on Red sign?\\[2pt]
\textbf{Locked Gold Answer}: The MUTCD does not specify this information. A correct response should state that the requested value or requirement cannot be determined from the manual rather than inventing an answer.\\[2pt]
\textbf{Locked Evidence}: Section 2B.60, Table 2B-1, PDF page 113.
}}%
\par\vspace{-\fboxrule}
\noindent\fcolorbox{gray!65!black}{gray!8!white}{%
\parbox{\dimexpr\linewidth-2\fboxsep-2\fboxrule\relax}{%
\small
\noindent\textbf{7. F026}\hfill\textit{Figure, Easy, Unanswerable, M-SDI 3/10, Absent Information}\\[2pt]
\textbf{Question}: What foundation depth is required for the sign posts drawn in Figure 2A-2?\\[2pt]
\textbf{Locked Gold Answer}: The MUTCD does not specify this information. A correct response should state that the requested value or requirement cannot be determined from the manual rather than inventing an answer.\\[2pt]
\textbf{Locked Evidence}: Sections 2A.13, 2A.18, Figure 2A-2, PDF page 92.
}}%
\par\vspace{-\fboxrule}
\noindent\fcolorbox{gray!65!black}{gray!8!white}{%
\parbox{\dimexpr\linewidth-2\fboxsep-2\fboxrule\relax}{%
\small
\noindent\textbf{8. F029}\hfill\textit{Figure, Easy, Unanswerable, M-SDI 3/10, Unsupported Precision}\\[2pt]
\textbf{Question}: What exact crossing angle, in degrees, is shown for the railroad tracks in Figure 8E-1?\\[2pt]
\textbf{Locked Gold Answer}: The MUTCD does not specify this information. A correct response should state that the requested value or requirement cannot be determined from the manual rather than inventing an answer.\\[2pt]
\textbf{Locked Evidence}: Sections 8E.01, 8E.02, Figure 8E-1, PDF page 1076.
}}%
\par\vspace{-\fboxrule}
\noindent\fcolorbox{gray!65!black}{gray!8!white}{%
\parbox{\dimexpr\linewidth-2\fboxsep-2\fboxrule\relax}{%
\small
\noindent\textbf{9. M028}\hfill\textit{Mixed, Medium, Unanswerable, M-SDI 6/10, Absent Information}\\[2pt]
\textbf{Question}: What were the sample size, participant demographics, and statistical significance values of the human-factors study underlying each treatment in Figure 2A-1?\\[2pt]
\textbf{Locked Gold Answer}: The MUTCD does not specify this information. A correct response should state that the requested value or requirement cannot be determined from the manual rather than inventing an answer.\\[2pt]
\textbf{Locked Evidence}: Section 2A.11, Figure 2A-1, PDF pages 89, 90.
}}%
\par\vspace{-\fboxrule}
\noindent\fcolorbox{gray!65!black}{gray!8!white}{%
\parbox{\dimexpr\linewidth-2\fboxsep-2\fboxrule\relax}{%
\small
\noindent\textbf{10. M030}\hfill\textit{Mixed, Hard, Unanswerable, M-SDI 7/10, Out Of Scope}\\[2pt]
\textbf{Question}: What exact dollar penalty and criminal classification apply when an agency violates a MUTCD Standard?\\[2pt]
\textbf{Locked Gold Answer}: The MUTCD does not specify this information. A correct response should state that the requested value or requirement cannot be determined from the manual rather than inventing an answer.\\[2pt]
\textbf{Locked Evidence}: Sections 1B.01, 1D.02, PDF pages 45, 72.
}}%
\par
\endgroup

\subsection{Easy Questions}

\begingroup
\setlength{\fboxsep}{5pt}
\noindent\fcolorbox{green!50!black}{green!50!black}{%
\parbox{\dimexpr\linewidth-2\fboxsep-2\fboxrule\relax}{%
\color{white}\centering\bfseries\strut Easy Questions: 10 Locked Examples%
}}%
\par\vspace{-\fboxrule}
\noindent\fcolorbox{green!50!black}{green!6!white}{%
\parbox{\dimexpr\linewidth-2\fboxsep-2\fboxrule\relax}{%
\small
\noindent\textbf{1. T001}\hfill\textit{Text, Easy, Answerable, M-SDI 2/10}\\[2pt]
\textbf{Question}: What is the purpose of the MUTCD, and what four objectives are used to achieve that purpose?\\[2pt]
\textbf{Locked Gold Answer}: The MUTCD establishes uniform national criteria for traffic control devices on streets, highways, pedestrian and bicycle facilities, and site roadways open to public travel. Its objectives are to promote safety, inclusion, and mobility; promote efficiency through national uniformity in device meaning and appearance; promote consistency in use, installation, and operation; and provide basic decision-making principles for traffic engineers.\\[2pt]
\textbf{Locked Evidence}: Section 1A.01, PDF page 42.
}}%
\par\vspace{-\fboxrule}
\noindent\fcolorbox{green!50!black}{green!6!white}{%
\parbox{\dimexpr\linewidth-2\fboxsep-2\fboxrule\relax}{%
\small
\noindent\textbf{2. T003}\hfill\textit{Text, Easy, Answerable, M-SDI 1/10}\\[2pt]
\textbf{Question}: Are islands, curbs, speed humps, raised roadway surfaces, and rumble strips considered traffic control devices under the MUTCD?\\[2pt]
\textbf{Locked Gold Answer}: No. Infrastructure elements that physically restrict travel paths or speeds, including islands, curbs, speed humps, other raised surfaces, and transverse or longitudinal rumble strips, are not traffic control devices.\\[2pt]
\textbf{Locked Evidence}: Section 1A.02, PDF page 42.
}}%
\par\vspace{-\fboxrule}
\noindent\fcolorbox{green!50!black}{green!6!white}{%
\parbox{\dimexpr\linewidth-2\fboxsep-2\fboxrule\relax}{%
\small
\noindent\textbf{3. T014}\hfill\textit{Text, Easy, Answerable, M-SDI 2/10}\\[2pt]
\textbf{Question}: What general meanings does the MUTCD assign to fluorescent pink, orange, and purple?\\[2pt]
\textbf{Locked Gold Answer}: Fluorescent pink is for incident management, orange is for temporary traffic control, and purple is restricted to vehicles with registered electronic toll collection accounts.\\[2pt]
\textbf{Locked Evidence}: Section 1D.05, PDF page 74.
}}%
\par\vspace{-\fboxrule}
\noindent\fcolorbox{green!50!black}{green!6!white}{%
\parbox{\dimexpr\linewidth-2\fboxsep-2\fboxrule\relax}{%
\small
\noindent\textbf{4. T019}\hfill\textit{Text, Easy, Answerable, M-SDI 2/10}\\[2pt]
\textbf{Question}: What are the minimum mounting heights for roadside signs in rural areas and in business, commercial, or residential areas where parking, bicyclist, or pedestrian activity is likely?\\[2pt]
\textbf{Locked Gold Answer}: The minimum is 5 feet in rural areas and 7 feet in business, commercial, or residential areas where such activity or obstruction is likely, measured as specified from the near pavement edge or curb.\\[2pt]
\textbf{Locked Evidence}: Section 2A.15, PDF page 99.
}}%
\par\vspace{-\fboxrule}
\noindent\fcolorbox{green!50!black}{green!6!white}{%
\parbox{\dimexpr\linewidth-2\fboxsep-2\fboxrule\relax}{%
\small
\noindent\textbf{5. TB011}\hfill\textit{Table, Easy, Answerable, M-SDI 3/10}\\[2pt]
\textbf{Question}: At a posted or 85th-percentile speed of 60 mph, what advance placement distance does Table 2C-3 show for deceleration to an advisory speed of 30 mph?\\[2pt]
\textbf{Locked Gold Answer}: 200 feet.\\[2pt]
\textbf{Locked Evidence}: Section 2C.04, Table 2C-3, PDF page 193.
}}%
\par\vspace{-\fboxrule}
\noindent\fcolorbox{green!50!black}{green!6!white}{%
\parbox{\dimexpr\linewidth-2\fboxsep-2\fboxrule\relax}{%
\small
\noindent\textbf{6. TB014}\hfill\textit{Table, Easy, Answerable, M-SDI 3/10}\\[2pt]
\textbf{Question}: What minimum passing sight distance does Table 3B-1 specify for a speed of 55 mph?\\[2pt]
\textbf{Locked Gold Answer}: 900 feet.\\[2pt]
\textbf{Locked Evidence}: Section 3B.03, Table 3B-1, PDF page 584.
}}%
\par\vspace{-\fboxrule}
\noindent\fcolorbox{green!50!black}{green!6!white}{%
\parbox{\dimexpr\linewidth-2\fboxsep-2\fboxrule\relax}{%
\small
\noindent\textbf{7. TB019}\hfill\textit{Table, Easy, Answerable, M-SDI 3/10}\\[2pt]
\textbf{Question}: What minimum sight distance for signal visibility does Table 4D-2 list at an 85th-percentile speed of 45 mph?\\[2pt]
\textbf{Locked Gold Answer}: 460 feet.\\[2pt]
\textbf{Locked Evidence}: Section 4D.06, Table 4D-2, PDF page 711.
}}%
\par\vspace{-\fboxrule}
\noindent\fcolorbox{green!50!black}{green!6!white}{%
\parbox{\dimexpr\linewidth-2\fboxsep-2\fboxrule\relax}{%
\small
\noindent\textbf{8. F001}\hfill\textit{Figure, Easy, Answerable, M-SDI 3/10}\\[2pt]
\textbf{Question}: In Figure 1B-1, what happens immediately after FHWA approves a request to experiment?\\[2pt]
\textbf{Locked Gold Answer}: The requesting jurisdiction installs the experimental traffic control device.\\[2pt]
\textbf{Locked Evidence}: Section 1B.05, Figure 1B-1, PDF page 48.
}}%
\par\vspace{-\fboxrule}
\noindent\fcolorbox{green!50!black}{green!6!white}{%
\parbox{\dimexpr\linewidth-2\fboxsep-2\fboxrule\relax}{%
\small
\noindent\textbf{9. F009}\hfill\textit{Figure, Easy, Answerable, M-SDI 3/10}\\[2pt]
\textbf{Question}: Which sign is placed above the other in every paired assembly shown in Figure 2B-17?\\[2pt]
\textbf{Locked Gold Answer}: The DO NOT ENTER sign is above the WRONG WAY sign.\\[2pt]
\textbf{Locked Evidence}: Section 2B.48, Figure 2B-17, PDF page 156.
}}%
\par\vspace{-\fboxrule}
\noindent\fcolorbox{green!50!black}{green!6!white}{%
\parbox{\dimexpr\linewidth-2\fboxsep-2\fboxrule\relax}{%
\small
\noindent\textbf{10. F024}\hfill\textit{Figure, Easy, Answerable, M-SDI 3/10}\\[2pt]
\textbf{Question}: In Figure 9E-12, what minimum distance separates the advance motor-vehicle stop line from the intersection stop line across the bicycle box?\\[2pt]
\textbf{Locked Gold Answer}: 10 feet minimum.\\[2pt]
\textbf{Locked Evidence}: Section 9E.12, Figure 9E-12, PDF page 1147.
}}%
\par
\endgroup

\subsection{Medium Questions}

\begingroup
\setlength{\fboxsep}{5pt}
\noindent\fcolorbox{orange!65!black}{orange!65!black}{%
\parbox{\dimexpr\linewidth-2\fboxsep-2\fboxrule\relax}{%
\color{white}\centering\bfseries\strut Medium Questions: 10 Locked Examples%
}}%
\par\vspace{-\fboxrule}
\noindent\fcolorbox{orange!65!black}{yellow!8!white}{%
\parbox{\dimexpr\linewidth-2\fboxsep-2\fboxrule\relax}{%
\small
\noindent\textbf{1. T009}\hfill\textit{Text, Medium, Answerable, M-SDI 4/10}\\[2pt]
\textbf{Question}: When a non-compliant traffic control device is replaced because it is damaged, missing, or no longer serviceable, what is the general rule and the limited replacement-in-kind exception?\\[2pt]
\textbf{Locked Gold Answer}: The replacement must generally be a compliant device. A non-compliant device may be replaced in kind only when engineering judgment finds that a lone compliant device among adjacent non-compliant devices would confuse users and/or the planned replacement schedule for the whole series will achieve timely compliance.\\[2pt]
\textbf{Locked Evidence}: Section 1B.03, PDF page 47.
}}%
\par\vspace{-\fboxrule}
\noindent\fcolorbox{orange!65!black}{yellow!8!white}{%
\parbox{\dimexpr\linewidth-2\fboxsep-2\fboxrule\relax}{%
\small
\noindent\textbf{2. T012}\hfill\textit{Text, Medium, Answerable, M-SDI 5/10}\\[2pt]
\textbf{Question}: What public-domain condition applies to a new traffic control device proposed for experimentation or interim approval?\\[2pt]
\textbf{Locked Gold Answer}: The traffic-control message, design, or application must be in the public domain and not protected by patent, trademark, or copyright. Legally binding statements are required from the relevant agency/developer/supplier for experimentation. Protection of internal components or manufacturing methods can still be allowed.\\[2pt]
\textbf{Locked Evidence}: Sections 1B.05, 1D.06, PDF pages 50, 74, 75.
}}%
\par\vspace{-\fboxrule}
\noindent\fcolorbox{orange!65!black}{yellow!8!white}{%
\parbox{\dimexpr\linewidth-2\fboxsep-2\fboxrule\relax}{%
\small
\noindent\textbf{3. T035}\hfill\textit{Text, Medium, Answerable, M-SDI 5/10}\\[2pt]
\textbf{Question}: Does satisfying a traffic signal warrant automatically require installation of a traffic control signal?\\[2pt]
\textbf{Locked Gold Answer}: No. A warrant is a threshold that triggers analysis; it is not conclusive justification. Installation must be based on an engineering study considering safety, operation, alternatives, and the site. A signal can also be installed where no warrant is met if an engineering study documents why it is the best solution.\\[2pt]
\textbf{Locked Evidence}: Sections 4B.04, 4C.01, PDF pages 691, 694.
}}%
\par\vspace{-\fboxrule}
\noindent\fcolorbox{orange!65!black}{yellow!8!white}{%
\parbox{\dimexpr\linewidth-2\fboxsep-2\fboxrule\relax}{%
\small
\noindent\textbf{4. T046}\hfill\textit{Text, Medium, Answerable, M-SDI 4/10}\\[2pt]
\textbf{Question}: When two-way traffic must share one lane for a limited distance, what coordination rule and control methods apply?\\[2pt]
\textbf{Locked Gold Answer}: Movements from both ends must be coordinated, except for a short, low-volume, mutually visible work zone that can self-regulate. Methods include flaggers, flag transfer, a pilot car, temporary traffic signals, or stop/yield control.\\[2pt]
\textbf{Locked Evidence}: Section 6E.01, PDF page 828.
}}%
\par\vspace{-\fboxrule}
\noindent\fcolorbox{orange!65!black}{yellow!8!white}{%
\parbox{\dimexpr\linewidth-2\fboxsep-2\fboxrule\relax}{%
\small
\noindent\textbf{5. TB003}\hfill\textit{Table, Medium, Answerable, M-SDI 4/10}\\[2pt]
\textbf{Question}: In Table 1D-2, how is 'Road Work' abbreviated on a portable changeable message sign, and what may follow it?\\[2pt]
\textbf{Locked Gold Answer}: RD WK; it may be followed by AHEAD or a distance, such as 'RD WK 1 MILE.'\\[2pt]
\textbf{Locked Evidence}: Section 1D.08, Table 1D-2, PDF page 77.
}}%
\par\vspace{-\fboxrule}
\noindent\fcolorbox{orange!65!black}{yellow!8!white}{%
\parbox{\dimexpr\linewidth-2\fboxsep-2\fboxrule\relax}{%
\small
\noindent\textbf{6. TB007}\hfill\textit{Table, Medium, Answerable, M-SDI 5/10}\\[2pt]
\textbf{Question}: For white-on-red signs, what minimum maintained retroreflectivity values and contrast criterion are shown in Table 2A-5?\\[2pt]
\textbf{Locked Gold Answer}: White at least 35 cd/lx/m$^2$, red at least 7 cd/lx/m$^2$, and a white-to-red contrast ratio of at least 3:1.\\[2pt]
\textbf{Locked Evidence}: Section 2A.22, Table 2A-5, PDF page 104.
}}%
\par\vspace{-\fboxrule}
\noindent\fcolorbox{orange!65!black}{yellow!8!white}{%
\parbox{\dimexpr\linewidth-2\fboxsep-2\fboxrule\relax}{%
\small
\noindent\textbf{7. TB018}\hfill\textit{Table, Medium, Answerable, M-SDI 4/10}\\[2pt]
\textbf{Question}: For a four-leg intersection, what one-year crash thresholds are listed in Table 4C-2 for all-severity angle/pedestrian crashes and for fatal-and-injury angle/pedestrian crashes?\\[2pt]
\textbf{Locked Gold Answer}: Five all-severity angle and pedestrian crashes, or three fatal-and-injury angle and pedestrian crashes.\\[2pt]
\textbf{Locked Evidence}: Section 4C.08, Table 4C-2, PDF page 702.
}}%
\par\vspace{-\fboxrule}
\noindent\fcolorbox{orange!65!black}{yellow!8!white}{%
\parbox{\dimexpr\linewidth-2\fboxsep-2\fboxrule\relax}{%
\small
\noindent\textbf{8. F003}\hfill\textit{Figure, Medium, Answerable, M-SDI 4/10}\\[2pt]
\textbf{Question}: What conspicuity treatment is labeled 'E' in Figure 2A-1?\\[2pt]
\textbf{Locked Gold Answer}: A vertical retroreflective strip on the sign support.\\[2pt]
\textbf{Locked Evidence}: Section 2A.11, Figure 2A-1, PDF page 89.
}}%
\par\vspace{-\fboxrule}
\noindent\fcolorbox{orange!65!black}{yellow!8!white}{%
\parbox{\dimexpr\linewidth-2\fboxsep-2\fboxrule\relax}{%
\small
\noindent\textbf{9. F017}\hfill\textit{Figure, Medium, Answerable, M-SDI 5/10}\\[2pt]
\textbf{Question}: What four main areas of a temporary traffic control zone are shown in Figure 6B-1, in order of travel?\\[2pt]
\textbf{Locked Gold Answer}: Advance warning area, transition area, activity area, and termination area.\\[2pt]
\textbf{Locked Evidence}: Section 6B.03-6B.07, Figure 6B-1, PDF page 813.
}}%
\par\vspace{-\fboxrule}
\noindent\fcolorbox{orange!65!black}{yellow!8!white}{%
\parbox{\dimexpr\linewidth-2\fboxsep-2\fboxrule\relax}{%
\small
\noindent\textbf{10. M004}\hfill\textit{Mixed, Medium, Answerable, M-SDI 6/10}\\[2pt]
\textbf{Question}: A portable CMS designer proposes 'DLY' to mean DELAY. Using Section 1D.08 and Table 1D-3, is this acceptable and why?\\[2pt]
\textbf{Locked Gold Answer}: No. DLY is explicitly prohibited because it can be misinterpreted as DAILY.\\[2pt]
\textbf{Locked Evidence}: Section 1D.08, Table 1D-3, PDF pages 75, 78.
}}%
\par
\endgroup

\subsection{Hard Questions}

\begingroup
\setlength{\fboxsep}{5pt}
\noindent\fcolorbox{red!60!black}{red!60!black}{%
\parbox{\dimexpr\linewidth-2\fboxsep-2\fboxrule\relax}{%
\color{white}\centering\bfseries\strut Hard Questions: 10 Locked Examples%
}}%
\par\vspace{-\fboxrule}
\noindent\fcolorbox{red!60!black}{red!5!white}{%
\parbox{\dimexpr\linewidth-2\fboxsep-2\fboxrule\relax}{%
\small
\noindent\textbf{1. M002}\hfill\textit{Mixed, Hard, Answerable, M-SDI 7/10}\\[2pt]
\textbf{Question}: Using Figure 2A-1 and Section 2A.11, describe the requirements for a retroreflective strip used on a sign support to enhance conspicuity.\\[2pt]
\textbf{Locked Gold Answer}: The strip must be at least 2 inches wide, extend from the sign down the full support to within 2 feet above the near edge of the roadway, match the sign background color except that YIELD and DO NOT ENTER strips are red, and contain no legend or other information.\\[2pt]
\textbf{Locked Evidence}: Section 2A.11, Figure 2A-1, PDF pages 89, 90.
}}%
\par\vspace{-\fboxrule}
\noindent\fcolorbox{red!60!black}{red!5!white}{%
\parbox{\dimexpr\linewidth-2\fboxsep-2\fboxrule\relax}{%
\small
\noindent\textbf{2. M003}\hfill\textit{Mixed, Hard, Answerable, M-SDI 8/10}\\[2pt]
\textbf{Question}: For a basic rural roadside sign, combine Figure 2A-2 with Sections 2A.15 and 2A.16 to state the normal minimum mounting height and lateral offset.\\[2pt]
\textbf{Locked Gold Answer}: The minimum mounting height is 5 feet. The normal minimum lateral offset is 12 feet from the edge of the traveled way, or 6 feet from the edge of a shoulder wider than 6 feet.\\[2pt]
\textbf{Locked Evidence}: Sections 2A.15, 2A.16, Figure 2A-2, PDF pages 92, 99, 101.
}}%
\par\vspace{-\fboxrule}
\noindent\fcolorbox{red!60!black}{red!5!white}{%
\parbox{\dimexpr\linewidth-2\fboxsep-2\fboxrule\relax}{%
\small
\noindent\textbf{3. M005}\hfill\textit{Mixed, Hard, Answerable, M-SDI 7/10}\\[2pt]
\textbf{Question}: At an all-way stop, what plaque arrangement and appearance are required by Section 2B.04 and Figure 2B-1, and may a 4-WAY plaque be substituted?\\[2pt]
\textbf{Locked Gold Answer}: An ALL-WAY (R1-3P) plaque must be mounted below each STOP sign. It has a white legend and border on a red background. A 4-WAY or other numeric-way plaque may not be substituted.\\[2pt]
\textbf{Locked Evidence}: Section 2B.04, Figure 2B-1, PDF page 115.
}}%
\par\vspace{-\fboxrule}
\noindent\fcolorbox{red!60!black}{red!5!white}{%
\parbox{\dimexpr\linewidth-2\fboxsep-2\fboxrule\relax}{%
\small
\noindent\textbf{4. M007}\hfill\textit{Mixed, Hard, Answerable, M-SDI 8/10}\\[2pt]
\textbf{Question}: If an agency is using an approved retroreflectivity-management method but a few individual signs temporarily fall below Table 2A-5, is the agency automatically out of compliance? Include the white-on-red thresholds relevant to STOP-type signs.\\[2pt]
\textbf{Locked Gold Answer}: Not automatically. The agency complies by having and using a method designed to maintain the minimums, even if some individual signs fall below them at a particular moment. For white on red, the table gives white at least 35, red at least 7 cd/lx/m$^2$, and at least a 3:1 white-to-red contrast ratio.\\[2pt]
\textbf{Locked Evidence}: Section 2A.22, Table 2A-5, PDF pages 103, 104.
}}%
\par\vspace{-\fboxrule}
\noindent\fcolorbox{red!60!black}{red!5!white}{%
\parbox{\dimexpr\linewidth-2\fboxsep-2\fboxrule\relax}{%
\small
\noindent\textbf{5. M008}\hfill\textit{Mixed, Hard, Answerable, M-SDI 8/10}\\[2pt]
\textbf{Question}: On a multi-lane roadway posted at 35 mph, may the single-lane regulatory-sign size be used, and what size must a STOP sign still use?\\[2pt]
\textbf{Locked Gold Answer}: For regulatory signs other than STOP, the single-lane size may be used at 35 mph or less. The STOP-sign exception remains: the multi-lane STOP size is 36 x 36 inches.\\[2pt]
\textbf{Locked Evidence}: Section 2B.03, Table 2B-1, PDF pages 108, 109.
}}%
\par\vspace{-\fboxrule}
\noindent\fcolorbox{red!60!black}{red!5!white}{%
\parbox{\dimexpr\linewidth-2\fboxsep-2\fboxrule\relax}{%
\small
\noindent\textbf{6. M010}\hfill\textit{Mixed, Hard, Answerable, M-SDI 8/10}\\[2pt]
\textbf{Question}: For a curve with an advisory speed of 40 mph and a radius within 401 to 700 feet, what Chevron Alignment sign spacing and placement are indicated?\\[2pt]
\textbf{Locked Gold Answer}: Typical spacing is 120 feet. The signs must be on the outside of the curve, approximately at right angles to approaching traffic, with a minimum mounting height of 4 feet.\\[2pt]
\textbf{Locked Evidence}: Section 2C.08, Table 2C-5, PDF page 199.
}}%
\par\vspace{-\fboxrule}
\noindent\fcolorbox{red!60!black}{red!5!white}{%
\parbox{\dimexpr\linewidth-2\fboxsep-2\fboxrule\relax}{%
\small
\noindent\textbf{7. M012}\hfill\textit{Mixed, Hard, Answerable, M-SDI 9/10}\\[2pt]
\textbf{Question}: For Warrant 1 Condition A with at least two major-street lanes and one minor-street lane, what volumes must exist and over what time basis, and does meeting them compel signal installation?\\[2pt]
\textbf{Locked Gold Answer}: For each of the same 8 selected hours, 600 vph must exist on the major street (both approaches total) and 150 vph on the critical minor approach (one direction) at the 100-percent thresholds. Meeting the warrant does not compel installation; an engineering study must determine whether a signal is justified.\\[2pt]
\textbf{Locked Evidence}: Sections 4C.01, 4C.02, Table 4C-1, PDF page 694.
}}%
\par\vspace{-\fboxrule}
\noindent\fcolorbox{red!60!black}{red!5!white}{%
\parbox{\dimexpr\linewidth-2\fboxsep-2\fboxrule\relax}{%
\small
\noindent\textbf{8. M015}\hfill\textit{Mixed, Hard, Answerable, M-SDI 7/10}\\[2pt]
\textbf{Question}: Using Section 4I.06 and Figure 4I-4, describe the pedestrian interval order, minimum buffer interval, and walking speed normally used for clearance timing.\\[2pt]
\textbf{Locked Gold Answer}: The order is WALK, flashing DONT WALK pedestrian change interval, then steady DONT WALK buffer interval. The buffer interval is at least 2 seconds before conflicting traffic is released, and clearance is normally evaluated using 3.5 feet per second.\\[2pt]
\textbf{Locked Evidence}: Section 4I.06, Figure 4I-4, PDF pages 764, 765.
}}%
\par\vspace{-\fboxrule}
\noindent\fcolorbox{red!60!black}{red!5!white}{%
\parbox{\dimexpr\linewidth-2\fboxsep-2\fboxrule\relax}{%
\small
\noindent\textbf{9. M021}\hfill\textit{Mixed, Hard, Answerable, M-SDI 7/10}\\[2pt]
\textbf{Question}: For a Type C arrow board in Figure 6L-3, state its minimum size, legibility distance, minimum number of elements, and required categories of display modes.\\[2pt]
\textbf{Locked Gold Answer}: Minimum size 96 x 48 inches, minimum legibility distance 1 mile, and at least 15 elements. It must provide at least one directional merge mode, the flashing double-arrow mode, and at least one caution mode.\\[2pt]
\textbf{Locked Evidence}: Section 6L.06, Figure 6L-3, PDF page 875.
}}%
\par\vspace{-\fboxrule}
\noindent\fcolorbox{red!60!black}{red!5!white}{%
\parbox{\dimexpr\linewidth-2\fboxsep-2\fboxrule\relax}{%
\small
\noindent\textbf{10. M024}\hfill\textit{Mixed, Hard, Answerable, M-SDI 7/10}\\[2pt]
\textbf{Question}: Using Section 9E.12 and Figure 9E-12, state the minimum bicycle-box length, required pavement symbol, turn-on-red rule, and green-pavement coverage rule.\\[2pt]
\textbf{Locked Gold Answer}: The bicycle box must be at least 10 feet long, include at least one bicycle symbol, prohibit turns on red from its lane, and, if green pavement is used, fill the entire limits of the box.\\[2pt]
\textbf{Locked Evidence}: Section 9E.12, Figure 9E-12, PDF pages 1146, 1147.
}}%
\par
\endgroup